\documentclass{article}

\usepackage{arxiv}

\usepackage[utf8]{inputenc} 
\usepackage[T1]{fontenc}    
\usepackage{hyperref}       
\usepackage{url}            
\usepackage{booktabs}       
\usepackage{amsfonts}       
\usepackage{nicefrac}       
\usepackage{microtype}      
\usepackage{lipsum}		
\usepackage{graphicx}
\usepackage{natbib}
\usepackage{doi}
\usepackage{amsmath}
\usepackage{tabularx}
\usepackage{multirow}
\usepackage{rotating}
\usepackage{subcaption}
\usepackage{blindtext}

\title{Explaining AutoClustering: Uncovering Meta-Feature Contribution in AutoML for Clustering}

\date{} 					

\author{ \href{https://orcid.org/0000-0000-0000-0000}{\includegraphics[scale=0.06]{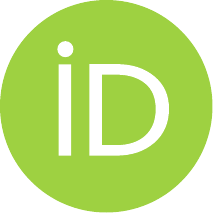}\hspace{1mm}Matheus Camilo da Silva} \\
	Department of Mathematics, Computer Science \\ and Geosciences\\
	University of Trieste\\
	Piazzale Europa, 1, 34127 Trieste TS, Italy. \\
	\texttt{matheus.camilodasilva@phd.units.it} \\
    \And \href{https://orcid.org/0000-0000-0000-0000}{\includegraphics[scale=0.06]{orcid.pdf}\hspace{1mm}Leonardo Arrighi} \\ Department of Mathematics, Computer Science \\ and Geosciences\\ University of Trieste\\ Piazzale Europa, 1, 34127 Trieste TS, Italy. \\ \texttt{leonardo.arrighi@phd.units.it} \\
    \And \href{https://orcid.org/0000-0000-0000-0000}{\includegraphics[scale=0.06]{orcid.pdf}\hspace{1mm}Ana Carolina Lorena} \\ Division of Computer Science\\ Aeronautics Institute of Technology\\ Praça Marechal Eduardo Gomes, 50, 12228-900 \\ São José dos Campos SP, Brazil. \\ \texttt{ana.lorena@gp.ita.br} \\
    \And \href{https://orcid.org/0000-0000-0000-0000}{\includegraphics[scale=0.06]{orcid.pdf}\hspace{1mm}Sylvio Barbon Junior} \\ Department of Mathematics, Computer Science \\ and Geosciences\\ University of Trieste\\ Piazzale Europa, 1, 34127 Trieste TS, Italy. \\ \texttt{sylvio.barbonjunior@units.it} \\
}

\renewcommand{\shorttitle}{Explaining AutoClustering}

\hypersetup{
pdftitle={Explaining AutoClustering: Uncovering Meta-Feature Contribution in AutoML for Clustering},
pdfsubject={q-bio.NC, q-bio.QM},
pdfauthor={David S.~Hippocampus, Elias D.~Striatum},
pdfkeywords={First keyword, Second keyword, More},
}

\begin{document}
\maketitle

\begin{abstract}
	AutoClustering methods aim to automate unsupervised learning tasks, including algorithm selection (AS), hyperparameter optimization (HPO), and pipeline synthesis (PS), by often leveraging meta-learning over dataset meta-features. While these systems often achieve strong performance, their recommendations are often difficult to justify: the influence of dataset meta-features on algorithm and hyperparameter choices is typically not exposed, limiting reliability, bias diagnostics, and efficient meta-feature engineering. This limits reliability and diagnostic insight for further improvements. In this work, we investigate the explainability of the meta-models in AutoClustering. We first review 22 existing methods and organize their meta-features into a structured taxonomy. We then apply a global explainability technique (i.e., Decision Predicate Graphs) to assess feature importance within meta-models from selected frameworks. Finally, we use local explainability tools such as SHAP (SHapley Additive exPlanations) to analyse specific clustering decisions. Our findings highlight consistent patterns in meta-feature relevance, identify structural weaknesses in current meta-learning strategies that can distort recommendations, and provide actionable guidance for more interpretable Automated Machine Learning (AutoML) design. This study therefore offers a practical foundation for increasing decision transparency in unsupervised learning automation.
\end{abstract}

\keywords{AutoML\and Clustering\and Explainable Artificial Intelligence\and XAI\and Meta-learning}

\section{Introduction}

Designing effective Machine Learning (ML) pipelines is a complex engineering task. Each stage, from data preprocessing and feature extraction to model selection and hyperparameter tuning, requires expert judgment and iterative experimentation~\citep{olson2016automatingbd, hutter2019automated}. Even minor design choices can substantially affect performance, making the process costly, time-consuming, and difficult to reproduce.

Automated Machine Learning (AutoML) emerged to alleviate this burden by automatically searching for and configuring ML pipelines~\citep{hutter2014, vanschoren2019meta}. AutoML frameworks typically decompose this automation into progressively complex subproblems: algorithm selection (AS), hyperparameter optimization (HPO), combined algorithm selection and hyperparameter optimization (CASH), pipeline synthesis (PS)~\citep{brazdil2022metalearning}, and in the context of deep learning, neural architecture search (NAS)~\citep{elsken2019nas}. While this hierarchy has driven major advances in supervised learning, extending such automation to unsupervised learning, particularly clustering, presents unique challenges~\citep{bahri2022automl}.

Unlike classification or regression, clustering lacks ground truth labels. This absence complicates the definition of objective functions, model evaluation, and therefore the automation of the entire pipeline~\citep{von2012clustering, halkidi2001clustering}. Conventional approaches rely on Cluster Validity Indices (CVIs), internal measures of data compactness and separation, to guide optimization~\citep{vendramin2010relative}. However, CVI-based optimization often encodes rigid assumptions about what constitutes a “good” clustering, which may not align with user intent or domain semantics. Moreover, many CVIs are inherently biased toward algorithms that optimize similar geometric or density-based criteria, creating circular preferences and limiting the diversity of solutions explored~\citep{gagolewski2021cluster}.

To address this limitation, recent work has explored meta-learning as a strategy for guiding AutoML in unsupervised settings. In this paradigm, knowledge from previously clustered datasets is reused to inform automation of configurations for new, unseen data~\citep{poulakis2024survey}. At the heart of this process are meta-features: numerical descriptors that summarize dataset characteristics ~\citep{kotlar2021, RIVOLLI2022108101, vanschoren2018meta}. These meta-features can serve as inputs to meta-models or surrogate models that predict the expected performance of clustering pipelines, enabling learned objectives that extend beyond static CVIs.

However, while extensive research has focused on designing and selecting meta-features, their influence on the meta-model’s recommendations is rarely analysed in a systematic way. Systematic here means: (a) explicit framework selection criteria; (b) unified meta-feature taxonomy across N frameworks; (c) identical explanation pipeline; (d) cross-framework comparability via normalized metrics. In most frameworks, the reasoning process by which a meta-model recommends one algorithm or configuration over another remains a “black box”~\citep{segel2023symbolic, moosbauer2022improving}, providing limited visibility into why a given algorithm or configuration is preferred. It is often unclear which meta-features drive these decisions, how strongly they contribute, or whether they encode biases toward specific dataset types. These biases may arise not only from dataset characteristics but also from the inductive biases of the algorithms themselves, which can shape the meta-model’s preferences in subtle and unintended ways.

This limited decision transparency creates important constraints for both practice and research. 
In practical deployments, stakeholders may be reluctant to rely on AutoClustering recommendations 
without access to the underlying rationale, particularly in high-stakes settings~\citep{moosbauer2023towards}. 
From an XAI perspective, explanations are needed not only to \emph{justify} recommendations, but also to 
\emph{support control} (e.g., auditing and intervention), \emph{enable improvement} (debugging and refinement of the meta-learning design), and \emph{facilitate discovery} of previously unnoticed regularities in the data or in the behaviour of candidate methods~\citep{adadi2018peeking}. 
On the research side, the absence of explainability hinders iterative development of meta-learning strategies, because it becomes difficult to diagnose 
why certain pipelines succeed or fail across dataset regimes~\citep{lindauer2024positionactionhumancenteredautoml, da2024benchmarking}. 
Moreover, when feature influence is not inspected, meta-models may overfit to spurious or weakly relevant dataset characteristics, reducing robustness and 
generalizability across diverse collections~\citep{mehrabi2021survey, pmlr-v81-buolamwini18a}.

To address these challenges, our work investigates how explainable AI (XAI) methods can be applied to uncover and interpret the influence of meta-features in AutoClustering systems. We present a systematic and multi-level investigation into the explainability of meta-models in AutoClustering. First, we review and analyse 22 existing frameworks, categorizing their meta-features into a unified taxonomy spanning six families: simple, statistical, information-theoretic, complexity-based, model-based, and landmarking features. Second, we apply the global explainability method DPG, proposed by ~\citet{arrighi2024decision} to reveal how meta-models combine these features into decision paths and thresholds. Finally, we complement this with local explainability via SHAP~\citep{lundberg2017unified}, to examine how individual meta-feature values influence particular AutoClustering recommendations.

Together, these components define the contributions of this work:

\begin{itemize}
    \item We provide a unified meta-feature taxonomy and a structured explainability study of AutoClustering, revealing recurrent drivers, systematic preference effects, and limitations to generalisation.
    \item A global explainability analysis using DPG to reveal decision structures, feature hierarchies, and salient interaction patterns in clustering meta-models.
    \item A local explainability analysis using SHAP to examine fine-grained instance-level feature attributions.
    \item Practical guidelines for building more transparent and robust AutoClustering systems, including recommendations for meta-feature selection and bias-aware meta-learning, based on combined global and local explanations
\end{itemize}

By exposing how meta-features are combined into decision rules and instance-level attributions, this work enables practitioners to justify and audit AutoClustering recommendations rather than treating them as unexpected outputs. At the same time, the observed concentration of influence and the ablation results show how explanations can be used to improve AutoClustering, reducing meta-feature extraction cost substantially while preserving most of the meta-model’s predictive power.

The remainder of the paper is organised as follows: Sections~\ref{sec:motiavation} and \ref{sec:background} introduce a motivating example and the required background; Sections~\ref{sec:related works} and \ref{sec:taxonomy} review related work and present our taxonomy of AutoClustering meta-features; Sections~\ref{sec:xai_mtl} and \ref{sec:xai_mtl} report global (DPG) and local (SHAP) explainability results and an explanation-driven ablation study; Section~\ref{sec:meta-model-ablation} discusses implications and limitations; and Section~\ref{sec:conclusion} concludes.

\section{Motivational Example}
\label{sec:motiavation}

To illustrate the importance of understanding the inference mechanisms behind AutoClustering meta-models, we present a scenario in which both global and local XAI tools reveal misleading meta-feature influence and guide refinement of clustering recommendations. Using the Problem-oriented AutoML in Clustering (PoAC) framework proposed by \citet{silva2024arxiv} to recommend a clustering pipeline, we focus on how meta-features influence algorithm selection and hyperparameter optimization, and how XAI techniques can identify meta-features that are weakly relevant or potentially misleading for the model.

Consider a new dataset $D_{\text{new}}$ representing consumer trajectories in a retail environment. The data exhibit noise and contain clusters with varied shapes and densities, characteristics that often challenge conventional clustering algorithms such as $k$-means. In such contexts, density-based methods like DBSCAN~\citep{ester1996density} are typically better suited, as they can identify arbitrarily shaped clusters and are inherently robust to noise~\citep{kriegel2011density}, provided that their density thresholds (e.g., \texttt{eps}, \texttt{min\_samples}) are well configured. This highlights the dual importance of algorithm choice and effective hyperparameter tuning, both of which are influenced by the meta-model’s interpretation of dataset characteristics.

Before presenting the extracted meta-features, we clarify that meta-feature categories correspond to the taxonomy of meta-features, grouping descriptors that capture similar types of dataset characteristics, such as statistical properties, information-theoretic properties, or performance-based landmarkers. Table~\ref{tab:mf_example} reports the numerical values of the meta-features extracted from $D_{\text{new}}$ by PoAC. These values are computed directly from the dataset and serve as inputs to the meta-model used for algorithm and hyperparameter recommendation.

\begin{center}
\label{tab:mf_example}
\begin{tabular}{l l l}
\toprule
\textbf{Meta-Feature} & \textbf{Category} & \textbf{Value} \\
\midrule
\texttt{hopkins} & Information-theoretic & 0.92 \\
\texttt{iq\_range.sd} & Statistical & 0.83 \\
\texttt{nr\_num} & Simple & 5.00 \\
\texttt{SIL} & Landmarker & -0.08 \\
\texttt{DBS} & Landmarker & 4.10 \\
\bottomrule
\end{tabular}
\end{center}

These values suggest a dataset with strong cluster tendency, as indicated by the Hopkins statistic (\texttt{hopkins} $\approx 1.00$), moderate dispersion (\texttt{iq\_range.sd} $> 0.60$), and poor performance of centroid-based clustering methods. The latter is reflected by a negative average SILhouette score (\texttt{SIL}), a landmarker that measures cluster compactness and separation, and by a high Davies--Bouldin score (\texttt{DBS}), which quantifies the ratio between intra-cluster similarity and inter-cluster separation. Together, these landmarkers indicate overlapping or non-spherical cluster structures that are typically challenging for convex clustering algorithms. Based on these meta-features, PoAC recommends DBSCAN with hyperparameters \texttt{eps} = 0.4 and \texttt{min\_samples} = 15.

To understand the rationale behind this recommendation, we apply the global explainability technique DPG. In DPG, predicates correspond to decision rules of the form ``meta-feature operator threshold'' (e.g., \texttt{hopkins} $>$ 0.85) and represent the structural components through which the meta-model encodes its global reasoning. Table~\ref{tab:dpg_predicates} lists the most influential predicates, ranked by Local Reaching Centrality (LRC), a measure of structural importance within the predicate graph.

\begin{center}
\label{tab:dpg_predicates}
\begin{tabular}{l l}
\toprule
\textbf{Predicate} & \textbf{LRC Score} \\
\midrule
\texttt{hopkins} $>$ 0.85 & 1.6600 \\
\texttt{iq\_range.sd} $>$ 0.6 & 1.5521 \\
\texttt{SIL} $\leq$ 0.0 & 1.3773 \\
\texttt{DBS} $>$ 3.0 & 1.2204 \\
\texttt{kurtosis.mean} $>$ 2.0 & 0.9596 \\
\bottomrule
\end{tabular}
\end{center}

The first three predicates align with expectations for density-based clustering, indicating high clusterability, data dispersion, and the unsuitability of centroid-based algorithms. However, \texttt{kurtosis.mean}, while recurrent in the learned DPG predicates, has limited interpretability in the context of spatial or density-based cluster separability. This is because kurtosis characterizes the tail-heaviness of feature distributions rather than geometric properties such as neighborhood density, local connectivity, or inter-point distances, which directly govern the behaviour of density-based clustering algorithms like DBSCAN.

To further examine this effect, we apply SHAP for local, instance-level analysis of the same recommendation. The SHAP explanation reveals the contribution of each meta-feature to the predicted suitability of DBSCAN for $D_{\text{new}}$, as shown in Table~\ref{tab:shap_values}.

\begin{center}
\label{tab:shap_values}
\begin{tabular}{l l l}
\toprule
\textbf{Meta-Feature} & \textbf{SHAP Value} & \textbf{Interpretation} \\
\midrule
\texttt{hopkins} & +0.24 & Strong positive contribution \\
\texttt{iq\_range.sd} & +0.17 & Supports preference for non-spherical clusters \\
\texttt{SIL} & +0.15 & Penalizes centroid-based methods \\
\texttt{kurtosis.mean} & +0.12 & Minor contribution \\
\texttt{nr\_num} & $-$0.06 & Slight penalty for low dimensionality \\
\bottomrule
\end{tabular}
\end{center}

Although \texttt{kurtosis.mean} appears structurally prominent in the global DPG analysis, its relatively small SHAP value indicates that it has limited influence on the specific recommendation for $D_{\text{new}}$. This discrepancy between global structural importance and local instance-level impact highlights the value of combining global and local explainability perspectives to uncover modeling biases and spurious feature dependencies.

Furthermore, the SHAP analysis provides interpretability for the selected hyperparameter \texttt{eps} = 0.4. The combination of high \texttt{hopkins} and moderate \texttt{iq\_range.sd} suggests well-separated clusters with variable local densities, for which smaller \texttt{eps} values are appropriate to avoid merging distinct dense regions. In contrast, larger \texttt{eps} values could over-smooth density differences and degrade clustering quality.

Overall, this example illustrates the practical role of explainability in AutoClustering: it helps surface meta-features that are structurally prominent yet weakly relevant in specific decisions (e.g., \texttt{kurtosis.mean}), provides case-level justification of recommendations through local attributions (SHAP), and links hyperparameter choices to interpretable signals in the meta-feature space. In doing so, it motivates our combined global--local analysis as a basis for more transparent, robust, and efficient unsupervised AutoML.

\section{Theoretical Background}
\label{sec:background}

This section outlines the foundational concepts underpinning our work. We begin with a detailed overview of meta-learning for AutoClustering, describing how past experience across datasets can be leveraged to recommend clustering pipelines in a data-driven, task-specific manner. We then discuss the role of explainability in AutoML systems, highlighting the challenges of interpreting meta-model decisions and the importance of transparent recommendation processes, particularly in unsupervised settings where ground-truth labels are unavailable.

\subsection{Meta-Learning for AutoClustering}
\label{sec:autoclustering_background}

Meta-learning, often referred to as “learning to learn”~\citep{vanschoren2019meta}, is a paradigm in which prior experience from solving multiple learning tasks is leveraged to improve performance on new, unseen ones~\citep{vilalta2002perspective, hospedales2021meta}. In this context, the term \textit{task} refers to a dataset along with its associated learning objective (e.g., classification, regression, or clustering), and should not be confused with AutoML tasks like AS or HPO, for instance. Rather than learning directly from raw data, meta-learning relies on \emph{meta-data}, such as past performance evaluations, dataset characteristics, or model behaviours, to drive generalization across tasks~\citep{brazdil2022metalearning}.

At the core of this process is the \emph{meta-learner}, which learns patterns in how pipelines perform across diverse tasks. This often involves training a \emph{meta-model}, that maps properties of tasks (e.g., meta-features) to properties of pipelines (e.g., expected performance). These meta-models can be used to predict performance~\citep{brazdil2022metalearning}, rank candidate pipelines~\citep{reif2012meta}, or recommend pipeline configurations~\citep{fusi2018probabilistic}, often in combination with different optimization or learning-to-rank strategies~\citep{vanschoren2019meta}.

In unsupervised learning, meta-learning can be more complex due to the absence of ground-truth labels and standard evaluation metrics like accuracy or recall. As a result, the meta-learner often rely on CVIs~\citep{tschechlov2023ml2dac, da2024benchmarking} or task-specific heuristics~\citep{bahri2022automl, chan2024autocd} as proxies for clustering quality. These proxies may be noisy or conflicting, introducing ambiguity and bias into the meta-learning process. Nevertheless, the meta-learning framework remains largely the same from supervised tasks. Based on two main phases, the offline (learning phase) and the online (inference phase). 

During the offline phase, a collection of historical datasets $\mathcal{D} = \{D_1, \ldots, D_n\}$ is processed to extract meta-features and evaluate clustering pipelines using internal or external validation measures. This results in a meta-dataset $\mathcal{D}_{\text{meta}}$, used to train a meta-model $\mathcal{M}: \mathcal{F} \mapsto \mathcal{Y}$, where $\mathcal{F} \in \mathbb{R}^d$ represents a vector of meta-features and $\mathcal{Y}$ denotes a clustering pipeline recommendation.

The choice of meta-model formulation, whether classification~\citep{lemke2015metalearning, tschechlov2023ml2dac}, regression~\citep{silva2024arxiv}, or learning-to-rank~\citep{desouto2008ranking, nascimento2009mining, reif2014automatic}, depends on the nature of the available tasks. Although AutoClustering aims to work in unsupervised settings, many meta-learning approaches rely on labeled datasets during the offline phase to compute external metrics like Adjusted Ranom Index (ARI) or Normalized Mutual Information, which provide more reliable performance signals. These signals enable classification or regression models. When labels are unavailable and only internal CVIs can be used, the problem is more often cast as a ranking task, where the goal is to order clustering pipelines by their expected relative performance~\cite{desouto2008ranking}.

The two main modelling approaches used during the offline phase, are:

\paragraph{(1) Ranking-Based Meta-Learning.}
This approach treats the meta-knowledge base as a repository of prior experience~\citep{GABBAY2021473}. The core assumption is that tasks with similar meta-features benefit from similar clustering pipelines. Given the meta-feature vector \( x_{\text{new}} \) of a new dataset, the system retrieves the most similar historical datasets \( \{x_i\} \) using a similarity function \( \text{sim}(\cdot, \cdot) \), and recommends the best-performing pipelines associated with them~\citep{de2008clustering,pimentel2018statistical, COHENSHAPIRA2021824, tschechlov2023ml2dac}.

Let \( \mathcal{P} = \{ \pi_1, \pi_2, \dots, \pi_n \} \) be the set of pipeline configurations evaluated during the offline phase, where each \( \pi_i \) denotes a clustering algorithm and its hyperparameters.

\begin{equation}
    \pi_{\text{rec}} = \arg\max_{\pi_i \in \mathcal{P}} \text{sim}(x_{\text{new}}, x_i),
\end{equation}

The top-\(K\) pipelines with the best historical performance among the most similar datasets are selected as:

\begin{equation}
    \{\pi_{(1)}, \dots, \pi_{(K)}\} = \text{TopK}_{\pi_i} \left( \text{perf}(\pi_i \mid \text{sim}(x_{\text{new}}, x_i)) \right).
\end{equation}

\paragraph{(2) Performance Prediction.}
The second approach is to train a meta-model \( f_{\text{meta}}: \mathcal{X} \rightarrow \mathbb{R} \) that directly estimates the performance of each pipeline based on the meta-features of the dataset~\citep{adam2015dealing,poulakis2020autoclust,silva2024arxiv}. Here, \( \mathcal{X} \subseteq \mathbb{R}^d \) represents the meta-feature space. For a new dataset \( \mathcal{D}_{\text{new}} \), with meta-features \( x_{\text{new}} \), the model predicts the performance of each candidate pipeline \( \pi_i \):

\begin{equation}
    \hat{y}_i = f_{\text{meta}}(x_{\text{new}}, \pi_i),
\end{equation}

and recommends the top-$K$ pipelines with the highest predicted scores:

\begin{equation}
    \{\pi_{(1)}, \dots, \pi_{(K)}\} = \text{TopK}_{\pi_i} \left( \hat{y}_i \right).
\end{equation}

In the online phase, a new dataset \( D_{\text{new}} \) is transformed into its meta-feature representation \( \mathcal{F}_{\text{new}} \) and passed to the meta-model \( \mathcal{M} \), which returns a clustering recommendation. This can support a variety of AutoML tasks, such as AS~\citep{desouto2008ranking, ferrari2012clustering}, HPO~\citep{poulakis2020autoclust, COHENSHAPIRA2021824}, CASH~\citep{tschechlov2023ml2dac}, or PS~\citep{elshawi2022tpe, silva2024arxiv}.

By leveraging prior knowledge, AutoClustering systems can avoid exhaustive searches and instead focus on the most promising regions of the pipeline space. This allows for efficient search and optimization using strategies such as Bayesian optimization, genetic algorithms~\citep{elshawi2022tpe, silva2024arxiv}, or regression-based ranking approaches~\citep{desouto2008ranking}, improving both performance and search efficiency in unsupervised settings.

Overall, the structure of AutoClustering systems underscores a central distinction between two meta-learning strategies. Ranking-based systems (Approach~1) primarily operate through similarity retrieval or relative preference estimation: rather than directly predicting absolute pipeline performance, the meta-learner infers rankings or similarities among pipelines based on their meta-feature proximity. While some ranking techniques may still learn an explicit scoring or preference function, the resulting decision process remains inherently relative and dependent on the choice of meta-features and similarity metrics. As a consequence, these systems can be robust when only internal CVIs are available, but they are highly sensitive to the representation of the meta-feature space and often struggle to generalize beyond regions well covered by previously observed datasets. By comparison, performance-prediction systems (Approach~2) learn an explicit function that maps meta-features to expected pipeline performance. This enables richer generalization and more nuanced reasoning about pipeline behaviour, but also introduces risks of overfitting to noisy or biased performance estimates, particularly when these are derived from imperfect internal CVIs.
 
Despite these differences, both paradigms share a critical dependency: \emph{all} decision-making occurs in the meta-feature space, without direct access to raw data or ground-truth labels. Consequently, applying XAI techniques becomes essential for understanding which meta-features drive recommendations, for enabling fair comparisons across systems, and for identifying costly or redundant features that contribute little to the final decision.

\subsection{Explainability and Automated Machine Learning}

XAI, in the context of AutoML, aims to make the behaviour of learning systems understandable by clarifying \emph{how} and \emph{why} models produce specific outputs~\citep{arrieta2020explainable,lundberg2017unified}. Beyond improving human understanding, XAI is increasingly framed as an enabling layer for (i) \emph{justifying} recommendations to stakeholders, (ii) \emph{supporting control} through auditing and intervention, (iii) \emph{improving} systems via debugging and refinement, and (iv) \emph{facilitating discovery} of previously unnoticed regularities or failure modes~\citep{adadi2018peeking,gunning2019xai,vilone2021notions,longo2024xai}. 

Although the terms \emph{interpretation} and \emph{explanation} are sometimes used interchangeably, recent work has emphasised useful distinctions. Following~\citet{lipton2018mythos}, we use \emph{interpretability} to denote the extent to which a human can reliably anticipate a model’s output given an input, and \emph{explainability} to denote the extent to which the internal mechanics or decision rationale can be articulated in human terms. In this view, interpretability is primarily associated with intrinsically transparent models (e.g., linear models or shallow decision trees), whereas explainability often relies on post-hoc methods that approximate or attribute the behaviour of complex models (e.g., ensembles or deep networks) without requiring them to be inherently transparent~\citep{lipton2018mythos,arrieta2020explainable}. 

Building on this distinction, XAI methods can be further differentiated by the \emph{scope} of the explanation they provide. \emph{Global} explainability characterises a model’s behaviour across the input space, for instance by identifying dominant features, decision rules, or recurring interaction patterns. By contrast, \emph{local} explainability focuses on individual predictions, attributing a specific output to particular input features for a given instance. Some approaches, such as SHAP, support both perspectives by producing instance-level attributions that can also be aggregated to summarise global trends.
nature~\citep{lundberg2017unified,ribeiro2016should}.

In AutoML, explanations are particularly valuable because the objects being optimised and recommended are not single predictions but \emph{pipeline decisions} (e.g., algorithm choice, validation criterion, or hyperparameter configuration). In AutoClustering, this decision-making occurs in a \emph{meta-feature space}, where each dataset is treated as an instance and the meta-model maps dataset descriptors to pipeline recommendations. This setting calls for complementary explanation views: a \emph{global} perspective that captures the dominant decision logic learned by the meta-model across datasets, and a \emph{local} perspective that justifies individual recommendations for a specific dataset. Accordingly, we adopt DPG as a global, model-agnostic technique that exposes rule-like decision structure and feature interactions, and SHAP as a local attribution method that quantifies how meta-feature values contribute to a particular recommendation. Together, DPG and SHAP provide a coherent explanation stack for AutoML systems, supporting both auditing of general decision patterns and case-by-case justification.

\subsubsection{Global Explainability of Decision Predicate Graphs}

Several global explainability methods have been proposed in recent years to understand the overall behaviour of machine learning models focusing on AutoML context. These include Partial Dependence Plots (PDPs)~\citep{moosbauer2021explaining} and symbolic regression~\citep{segel2023symbolic}. These techniques offer valuable insights on traditional AutoML, but they often struggle to preserve both interpretability and structural fidelity, especially when applied to meta-models operating in high-level feature spaces, such as those found in AutoClustering. Given the unique characteristics of meta-datasets used in AutoClustering approaches, we consider DPG to be particularly well-suited for providing global explanations.

DPG are a model-agnostic technique designed to produce global, interpretable summaries of machine learning models~\citep{arrighi2024decision,ceschin2025extending}. Rather than analysing raw predictions, DPG extracts symbolic decision rules, called \textit{predicates}, from trained models and connects them based on their influence and generality.

Given a set of input features $x \in \mathbb{R}^d$ and a trained model $f: \mathbb{R}^d \rightarrow \mathbb{R}$, DPG constructs a graph where each node represents a predicate (e.g., $x_j > \theta$), and edges represent logical pathways in the model’s decision space. The influence of each predicate is quantified using the LRC, which measures its reach and discriminative contribution.

\[
\text{LRC}(p_i) = \sum_{x \in \mathcal{D}} \mathbf{1}\left[p_i \text{ is satisfied in the path to } f(x)\right]
\]

This approach provides global insights such as: “datasets with high \texttt{hopkins} and low \texttt{SILhouette} scores tend to receive DBSCAN”. The symbolic structure and quantifiable relevance of DPG predicates make them well-suited to analyse the behaviour of different AutoClustering meta-models operating in the meta-feature space.

\vspace{0.5em}
\subsubsection{Local Explainability using SHAP}

Local explainability methods provide instance-level interpretations by attributing the model’s output to its input features. Among the most widely adopted methods is SHAP~\citep{lundberg2017unified}, which computes the contribution of each feature to a specific prediction using cooperative game theory. For a given prediction $f(x)$, the SHAP value $\phi_j$ for feature $j$ satisfies:

\[
f(x) = \phi_0 + \sum_{j=1}^d \phi_j, \quad \text{where } \phi_j \text{ reflects the marginal contribution of feature } j.
\]

In supervised learning, SHAP reveals which features most influence a prediction. In unsupervised or meta-learning contexts, this becomes more nuanced: predictions are not labels but cluster validity estimates or rankings, and features are not original dataset attributes, but meta-features summarizing dataset properties (e.g., \texttt{entropy}, \texttt{iq\_range.sd}, \texttt{hopkins}).

The main challenge in applying local explainability to AutoClustering methods is the abstraction of the input space, rather than interpreting a prediction based on raw features (e.g., pixel intensities or gene expressions), we must interpret predictions over meta-features, whose meaning is already a second-order abstraction of dataset behaviour.

Despite these challenges, local methods remain critical for justifying specific AutoClustering recommendations (e.g., “why DBSCAN with \texttt{eps} = 0.4 was selected”), especially in domains where transparency and trust are essential.

\section{Related Works}
\label{sec:related works}

Explainability in AutoML has gained growing attention as researchers seek to improve fairness, transparency, and trust in automated machine learning pipelines~\citep{lindauer2024positionactionhumancenteredautoml}. Much of this research has focused on supervised learning settings~\citep{van2018hyperparameter, garouani2022towards, moosbauer2023towards}, where labeled data allows for clearer evaluation and interpretability of model behaviour, optimization dynamics, and configuration choices.

A major line of research in this area is Hyperparameter Importance (HPI) analysis, which seeks to determine which hyperparameters most significantly influence model performance. A foundational method is f-ANOVA~\citep{hutter2014}, which uses variance decomposition to estimate the marginal contribution of each hyperparameter and their interactions, often revealing that a small subset of hyperparameters accounts for the majority of performance variation. Building on this, \citet{van2018hyperparameter} introduced a meta-learning framework that aggregates experimental results across many datasets to identify, for a given algorithm, which hyperparameters are generally most important and what typical good values are. In the same vein, \citet{probst2019} proposed measures for \textit{tunability} to identify which hyperparameters are worth tuning. 

Subsequent work has expanded these approaches. \citet{moosbauer2022improving} incorporated Bayesian optimization for probabilistic importance estimates, while \citet{jin2022} proposed efficient subsampling strategies to improve HPI estimation. \citet{watanabe2023} introduced PED-ANOVA, focusing on top-performing configurations, and \citet{theodorakopoulos2024} extended HPI to multi-objective settings, showing that hyperparameter impact can differ across goals such as accuracy, efficiency, and fairness.

In parallel, other explainability methods in AutoML have explored visualization and symbolic analysis. \citet{moosbauer2022} proposed PDPs with uncertainty estimates to interpret model responses to hyperparameter changes. \citet{segel2023symbolic} used symbolic regression to derive human-readable expressions linking hyperparameters to performance, aiding interpretability of the HPO landscape. Additional efforts like AutoAblation~\citep{sheikholeslami2021} automate ablation studies to understand pipeline components, while tools like CAVE~\citep{biedenkapp2019} visualize AutoML optimization trajectories and search behaviour.

Despite these advances, explainability in AutoClustering remains relatively unexplored. Clustering presents distinct challenges: it lacks ground truth labels and depends on subjective or domain-specific evaluation metrics~\citep{von2012clustering}, complicating both the automation and interpretability of AutoML task results. Moreover, many of the explainability techniques developed for supervised learning or HPO do not directly transfer to clustering-based AutoML systems.

Most AutoClustering frameworks rely on meta-learning, where historical performance on previous datasets guides algorithm selection for new, unseen data. To achieve this, meta-features are extracted to characterize datasets in a way that is predictive of clustering performance ~\citep{RIVOLLI2022108101, kotlar2021, brazdil2022metalearning}. However, meta-feature selection is often ad hoc or empirically driven, with limited standardization across frameworks. While some surveys, such as \citet{poulakis2024survey}, have catalogued the general use of meta-features, the field still lacks a systematic taxonomy comparing and organizing them across methods. This hinders comparative evaluation and makes it difficult to assess the generalizability or redundancy of feature sets used in different AutoClustering approaches.

\subsection{Positioning and Contribution of This Work}
\label{sec:rw_positioning}

Existing work in AutoClustering has largely emphasised predictive performance and the proposal of new meta-features or meta-learning strategies~\citep{RIVOLLI2022108101,kotlar2021,brazdil2022metalearning}, whereas comparative organisation of meta-features across frameworks remains limited to broad surveys~\citep{poulakis2024survey}. Moreover, although a few works in AutoML discuss transparency and user control at a high level~\citep{lindauer2024positionactionhumancenteredautoml}, there is still a lack of systematic evidence on how meta-features influence AutoClustering meta-models in practice, and how such evidence can be used to improve system design.

This paper addresses these gaps by bringing explainability to the \emph{meta-learning layer} of AutoClustering. First, we provide a unified taxonomy of meta-features used in 22 AutoClustering frameworks, supporting comparability across methods and highlighting redundancy and fragmentation in current feature design~\citep{poulakis2024survey}. Second, we combine complementary global and local explanation perspectives to analyse reconstructed meta-models from representative frameworks: global explanations via DPG expose decision structures and feature hierarchies, while local attributions via SHAP explain individual recommendations in the meta-feature space. Finally, we demonstrate that explanation evidence is not merely descriptive: it can be operationalised to guide meta-feature ablation, quantifying efficiency--accuracy trade-offs and motivating more transparent and cost-aware AutoClustering design. Together, these contributions position our work as a systematic step towards decision-transparent unsupervised AutoML, where recommendations can be inspected, audited, and refined rather than treated as opaque outputs.

\section{Meta-Feature Taxonomy for AutoClustering Meta-Learning} \label{sec:taxonomy}
In the context of AutoClustering, the domain of meta-features remains fragmented across studies, with varying definitions, computation costs, and empirical relevance. To analyse and explain AutoClustering meta-models in a comparable way across frameworks, we first need a shared view of the meta-feature space—what descriptors are used, how they relate, and how consistently they appear in the literature. 

We focused on works that explicitly define and utilize meta-features. Our selection spans from 2008 (the year of the first known study in this domain of~\citet{de2008clustering}) to 2025, and includes 22 representative frameworks chosen for their relevance and influence in the field, as shown in \autoref{tab:framework_overview}. These range from early contributions by \citet{desouto2008ranking}, \citet{nascimento2009mining}, and \citet{soares2009analysis}, which introduced meta-features in high-dimensional biological contexts, to more recent systems such as \citet{tschechlov2023ml2dac}, \citet{silva2024arxiv}, and \citet{fernandes2021towards}, which leverage synthetic datasets and classic benchmarks like UCI. This curated selection captures the evolution of meta-feature design across time and application domains, enabling a robust comparative taxonomy.

\begin{table}[!htb]
\caption{Overview of 22 AutoClustering frameworks reviewed in this study.}
\footnotesize
\setlength{\tabcolsep}{4pt}
\renewcommand{\arraystretch}{1.05}
\centering
\begin{tabularx}{\linewidth}{Xl}
\toprule
\textbf{Title} & \textbf{Reference} \\
\midrule
Ranking and selecting clustering algorithms using a meta-learning approach & \cite{desouto2008ranking} \\
Mining rules for the automatic selection process of clustering methods applied to cancer gene expression data & \cite{nascimento2009mining} \\
An analysis of meta-learning techniques for ranking clustering algorithms applied to artificial data & \cite{soares2009analysis} \\
Clustering algorithm recommendation: a meta-learning approach & \cite{ferrari2012clustering} \\
Dealing with overlapping clustering: a constraint-based approach to algorithm selection & \cite{adam2015dealing} \\
Clustering algorithm selection by meta-learning systems: a new distance-based problem characterization and ranking combination methods & \cite{ferrari2015clustering} \\
Extending meta-learning framework for clustering gene expression data with component-based algorithm design and internal evaluation measures & \cite{vukicevic2016extending} \\
Constraint-based clustering selection & \cite{van2017constraint} \\
Statistical versus distance-based meta-features for clustering algorithm recommendation using meta-learning & \cite{pimentel2018statistical} \\
A new data characterization for selecting clustering algorithms using meta-learning & \cite{pimentel2019new} \\
Unsupervised meta-learning for clustering algorithm recommendation & \cite{pimentel2019unsupervised} \\
AutoClust: a framework for automated clustering based on cluster validity indices & \cite{poulakis2020autoclust} \\
Automatic selection of clustering algorithms using supervised graph embedding & \cite{COHENSHAPIRA2021824} \\
cSmartML: a meta-learning-based framework for automated selection and hyperparameter tuning for clustering & \cite{elshawi2021csmartml} \\
Towards understanding clustering problems and algorithms: an instance space analysis & \cite{fernandes2021towards} \\
Isolation forests and landmarking-based representations for clustering algorithm recommendation using meta-learning & \cite{gabbay2021isolation} \\
AutoCluster: meta-learning-based ensemble method for automated unsupervised clustering & \cite{liu2021autocluster} \\
AutoML4Clust: efficient AutoML for clustering analyses & \cite{tschechlov2021automl4clust} \\
cSmartML-Glassbox: increasing transparency and controllability in automated clustering & \cite{elshawi2022csmartml} \\
TPE-AutoClust: a tree-based pipeline ensemble framework for automated clustering & \cite{elshawi2022tpe} \\
ML2DAC: meta-learning to democratize AutoML for clustering analyses & \cite{treder2023ml2dac} \\
Problem-oriented AutoML in clustering & \cite{silva2024arxiv} \\
\bottomrule
\end{tabularx}
\label{tab:framework_overview}
\end{table}

To better understand the design space of meta-feature usage in AutoClustering, we organize our taxonomy around two core components: the base datasets used to construct meta-datasets, and the types of meta-features extracted from them. The first component, Dataset Analysis, examines the sources, diversity, and characteristics of the datasets that serve as input for generating meta-features and training meta-models. The second, Meta-feature Categories, explores the different groups of meta-features adopted across frameworks, highlighting both widely used descriptors and more specialized or novel types. Together, these dimensions provide insight into how AutoClustering systems are grounded in data and guided by metadata to support the training and generalization of meta-models.

\subsection{Base Dataset Analysis}
The first step in training a meta-model for AutoClustering is the selection of base datasets that are representative of the kinds of clustering tasks the practitioner aims to support. This step is especially critical in the unsupervised setting, where there is no universally correct clustering, and both the data characteristics and the intended grouping logic must align with the target use cases~\citep{von2012clustering}. In this context, not only should the chosen datasets resemble those that the user expects to encounter, but the way they are labelled or clustered must also reflect the user’s intent on how new data should be clustered.

Its from these base datasets that meta-features are extracted, which ultimately enables the training of meta-models. In this sense, their properties such as domain, dimensionality and size, directly influence the kinds of patterns a meta-model can learn and generalize to new data. Therefore, understanding the composition and origin of these datasets is crucial to interpreting the design choices made across different frameworks.

Based on the selected works presented in \autoref{tab:framework_overview}, we identify four main groups of datasets commonly used in the literature:

\paragraph{Cancer Gene Expression Datasets.}
This dataset group consists of 35 publicly available microarray datasets curated and distributed by ~\cite{de2008clustering}. These datasets represent various types of cancer and are characterized by extremely high dimensionality (typically over 1,300 features) and very small sample sizes (fewer than 100 instances per dataset), making them particularly challenging for clustering tasks. The data originates from two main microarray platforms, Affymetrix and cDNA, which differ in measurement methodology. 

This group was predominantly used by earlier frameworks such as \cite{desouto2008ranking}, \cite{nascimento2009mining}, and \cite{vukicevic2016extending}, which aimed to evaluate clustering performance in biomedical domains where ground-truth labels (e.g., cancer subtypes) are available and well-studied. The small sample size and high dimensionality of these datasets present unique challenges, making them a natural proving ground for robustness to overfitting and noise. These datasets are less common in more recent work, likely due to their limited generalizability to other application domains and the increasing availability of more scalable benchmarking resources.

\paragraph{General-Purpose Benchmark Repositories.}
Datasets from the UCI Machine Learning Repository~\citep{kelly2023uci} and OpenML~\citep{OpenML2013} platform are frequently used in AutoClustering research due to their broad accessibility, domain diversity, and historical importance in benchmarking machine learning systems. These repositories offer a diverse collection of real-world datasets across various domains, including healthcare, finance, biology, and sensor data. They are widely used for benchmarking due to their accessibility, varying levels of complexity, and annotated ground truth labels. Datasets from these sources vary significantly in size, dimensionality, and cluster structure, making them suitable for evaluating the generalization ability of AutoClustering systems across heterogeneous settings.

Frameworks such as \cite{ferrari2015clustering}, \cite{COHENSHAPIRA2021824}, and \cite{pimentel2019new} make extensive use of datasets from UCI and OpenML, reflecting a shift toward real-world applicability and diversity. These repositories provide broad coverage of domains and data modalities, which helps assess the generalization capabilities of AutoClustering systems. Importantly, the inclusion of widely recognized benchmark datasets enhances reproducibility and comparability with prior work. The increased reliance on these repositories in recent literature highlights the community’s move toward more representative and accessible benchmarking standards.

\paragraph{Synthetic Data.}
This group consists of datasets that are algorithmically generated with the goal of simulating simplified yet plausible data distributions that may resemble real-world clustering tasks. By systematically varying parameters such as the number of clusters, dimensionality, sample size, or noise levels, synthetic datasets enable controlled experimentation and fine-grained benchmarking across a range of data morphologies. One influential early toolset is that of~\cite{handl2005cluster}, which introduced both Gaussian and ellipsoidal cluster generators, offering parametric control over shape, orientation, and overlap. More recent frameworks have made extensive use of the scikit-learn generators such as make\_blobs, make\_circles, and make\_moons, which allow researchers to generate synthetic data with varied geometric and topological properties using simple APIs. Additionally, packages like pymfe support mixture model-based data synthesis for high-throughput generation of clustering scenarios. 

These datasets are particularly useful for training meta-models or evaluating generalization performance, as they can be generated in large numbers while retaining controllable structural complexity. Frameworks such as \cite{soares2009analysis}, \cite{fernandes2021towards}, and \cite{treder2023ml2dac} have relied on synthetic data to explore performance under scalable and diverse conditions.

\paragraph{Clustering Benchmark Suites.}
In contrast, clustering benchmark suites such as SIPU~\citep{ClusteringDatasets} and Clustering-benchmark~\citep{barton2015clustering}, include carefully crafted datasets (e.g., TwoDiamonds, Jain, Smile) that are synthetically generated but not intended to simulate real-world data. Instead, they are designed explicitly to stress-test clustering algorithms on edge cases. These datasets often contain geometric configurations, overlapping boundaries, or chaining effects that deliberately challenge specific algorithmic assumptions, such as Euclidean distance-based similarity. For example, the TwoDiamonds dataset consists of two adjacent diamond-shaped clusters where points are uniformly distributed inside each shape, forming a continuous structure that is difficult to separate using traditional distance metrics~\citep{ultsch2003u}. These benchmark suites are used in works such as \cite{van2017constraint} and \cite{fernandes2021towards} typically to probe algorithmic robustness and sensitivity under known but difficult clustering scenarios.

\paragraph{}
Table~\ref{tab:dataset_summary} provides a detailed overview of dataset usage across the reviewed AutoClustering frameworks, sorted chronologically by publication year. For each framework, we report the number of datasets used, along with per-framework average values for the number of samples, dimensions, and ground-truth clusters across the datasets considered (when available). The dataset groups (A–D) correspond to the categories introduced above.

\begin{table}[ht!]
\caption{Summary of dataset usage across AutoClustering frameworks. 
Dataset groups:
\textbf{A} = Cancer gene expression microarray data, 
\textbf{B} = Synthetic datasets, 
\textbf{C} = General-purpose benchmark repositories, 
\textbf{D} = Clustering benchmarking collections. 
\textit{Note:} Numerical columns for samples, dimensions, and clusters report \emph{average values per framework} across the datasets used. Parentheses indicate different dataset settings within the same publication.}

\scriptsize  
\setlength{\tabcolsep}{3pt} 
\renewcommand{\arraystretch}{1.1}
\centering
\begin{tabularx}{\linewidth}{
  l   
  c   
  >{\raggedleft\arraybackslash}X  
  >{\raggedleft\arraybackslash}X  
  >{\raggedleft\arraybackslash}X  
  >{\raggedleft\arraybackslash}X  
}
\toprule
\textbf{Framework} & \textbf{Group} & \textbf{\#Datasets} & \textbf{\#Samples} & \textbf{\#Dimensions} & \textbf{\#Clusters} \\
\midrule
\cite{desouto2008ranking} & A & 32 & 76.58 & 1383.38 & 3.00 \\
\cite{nascimento2009mining} & A & 35 & 76.58 & 1383.38 & 3.00 \\
\cite{soares2009analysis} & B & 160 & 309.40 & 75.00 & 7.50 \\
\cite{ferrari2012clustering} & C & 30 & 1636.23 & 36.20 & 7.50 \\
\cite{adam2015dealing} (1) & C & 14 & 522.93 & 11.07 & 3.92 \\
\cite{adam2015dealing} (2) & -- & 22 & -- & -- & -- \\
\cite{ferrari2015clustering} & C & 84 & 369.00 & 87.22 & 9.45 \\
\cite{vukicevic2016extending} & A & 30 & 89.77 & 1713.67 & 3.70 \\
\cite{van2017constraint} (1) & C & 16 & 804.62 & 273.12 & 4.69 \\
\cite{van2017constraint} (2) & D & 5 & 1954.40 & 2.00 & 3.40 \\
\cite{pimentel2018statistical} & -- & 218 & -- & -- & -- \\
\cite{pimentel2019new} & C & 219 & 4799.85 & 19.96 & 5.84 \\
\cite{pimentel2019unsupervised} & -- & 57 & -- & -- & -- \\
\cite{poulakis2020autoclust} & C & 24 & 810.04 & 27.42 & 4.54 \\
\cite{COHENSHAPIRA2021824} & C & 210 & 726.79 & 22.29 & 165.46 \\
\cite{elshawi2021csmartml} (1) & C & 12 & 922857.04 & 48.50 & 5.50 \\
\cite{elshawi2021csmartml} (2) & D & 15 & 1950.13 & 8.13 & 13.00 \\
\cite{fernandes2021towards} (1) & B & 160 & 1859.25 & 40.50 & 18.50 \\
\cite{fernandes2021towards} (2) & C & 219 & 275.27 & 16.73 & 23.62 \\
\cite{fernandes2021towards} (3) & D & 220 & 3466.69 & 191.21 & 8.53 \\
\cite{gabbay2021isolation} & C & 100 & 2259.55 & 58.85 & 237.07 \\
\cite{liu2021autocluster} & -- & 150 & -- & -- & -- \\
\cite{tschechlov2021automl4clust} & C & 5 & 8691.80 & 142.40 & 12.00 \\
\cite{elshawi2022csmartml} & -- & 200 & -- & -- & -- \\
\cite{elshawi2022tpe} & -- & 118 & -- & -- & -- \\
\cite{treder2023ml2dac} & B & 78 & 5333.33 & 21.38 & 21.38 \\
\cite{silva2024arxiv} & B & 6130 & 2081.86 & 41.55 & 16.54 \\
\bottomrule
\end{tabularx}

\label{tab:dataset_summary}
\end{table}

The composition of datasets used by AutoClustering frameworks has evolved substantially. Earlier works tended to rely on smaller, high-dimensional biomedical datasets (e.g., microarrays), whereas more recent frameworks have increasingly employed synthetic generators and large-scale general-purpose repositories like UCI and OpenML. This diversification reflects both an expansion in available data resources and a growing interest in benchmarking across varied domains. However, a notable issue persists: several frameworks did not provide detailed dataset characteristics (e.g., number of instances, features, or clusters), or failed to make their datasets publicly accessible. This lack of transparency poses challenges to reproducibility and comparability, undermining efforts to evaluate progress in the field rigorously.

\subsection{Meta-Feature Families}
Meta-features are quantitative descriptors computed from base datasets that characterize their properties in a way that facilitates the learning of patterns by meta-models regarding algorithm performance or suitability~\cite{castiello2005meta}. In the context of AutoClustering, meta-features are designed to capture not only structural, statistical, and geometric aspects of the input data, but also characteristics of the clustering partitions derived from them~\cite{de2008clustering, pimentel2019new}. These descriptors form the basis of the meta-space, enabling the comparison of tasks, the modeling of task similarity, and ultimately the prediction of effective clustering configurations.

The extraction of meta-features is a multi-step process involving careful selection, computation, and preprocessing. As summarized in~\citet{vanschoren2018meta}, many meta-features are derived from individual data attributes (e.g., feature-wise skewness) or combinations thereof (e.g., correlation matrices), requiring aggregation via summary statistics such as mean, standard deviation, and quartiles. Task-level normalization is often necessary to make these descriptors comparable across datasets~\citep{rivolli2018towards, rivolli2022meta}. The particular choices in which meta-features to extract and their preprocessing can significantly impact the quality of the resulting meta-models~\citep{pinto2016towards}.

Given the flexibility of the meta-feature extraction process, a vast number of possible descriptors can be constructed. The optimal set of meta-features often depends on the intent behind the meta-model: for instance, whether it aims to rank a clustering algorithm, tune hyperparameters, or recommend clustering pipeline configurations. As such, meta-feature selection should be informed by the specific downstream task, dataset domain and computational constraints~\citep{rivolli2018towards, reif2014automatic, castiello2005meta}.

To support interpretability and reuse, meta-features are frequently grouped into high-level families based on the type of information they encode. While multiple taxonomies exist in the literature~\citep{castiello2005meta, reif2014automatic, rivolli2018towards}, the six families outlined below cover the most commonly used descriptors across AutoClustering frameworks analysed in this work.

\paragraph{Simple/General.} This family of meta-features, is composed of basic univariate statistics computed over the input features, such as mean, variance, minimum, and maximum. These provide coarse yet computationally efficient summaries of the data distribution and are especially useful in high-dimensional or resource-constrained scenarios. They were employed in early AutoClustering efforts such as~\citet{desouto2008ranking, nascimento2009mining}, and remain widely adopted in more recent work~\citep{gabbay2021isolation,treder2023ml2dac,silva2024arxiv} due to their general applicability~\cite{liu2021autocluster}.

\paragraph{Statistical.} Offers a more nuanced view of dataset structure by capturing relationships between features or distributional shape. Common examples include skewness, kurtosis, pairwise correlation, and covariance. These descriptors are useful for identifying feature redundancy, symmetry, and marginal dependencies. They are mathematically straightforward and interpretable~\citep{castiello2005meta}, which makes them a staple in most of the AutoClustering frameworks.

\paragraph{Information-theoretical.} Quantifies the amount of information or uncertainty associated with the features or feature combinations. Measures such as entropy, mutual information, and signal-to-noise ratios fall under this category~\citep{castiello2005meta}. They are particularly appropriate to describe categorical attributes, but can also represent continuous values. In AutoClustering, ~\citep{ferrari2012clustering} were among the first to incorporate this family, drawing directly on the StatLog~\citep{michie1995machine} and METAL~\citep{kalousis2000model} meta-feature sets. Later on, ~\citep{pimentel2019unsupervised} expanded the original meta-features used in ~\citep{desouto2008ranking} by adding more meta-features families, including a number of information-theoretical descriptors. Other works, such as ~\citep{vukicevic2016extending} and ~\citep{silva2024arxiv}, have continued to leverage and refine these features, demonstrating their ongoing relevance for characterizing dataset complexity in unsupervised settings.

\paragraph{Landmarker.} Originally proposed by~\citet{pfahringer2000meta} in supervised learning, landmarkers characterize datasets by measuring the behaviour of simple, fast models. In unsupervised settings, where accuracy-based metrics are unavailable, this idea is adapted by running lightweight clustering algorithms and evaluating their partitions using CVIs such as SIL, DBS, or CHS. These meta-features capture how different clustering biases interact with the data, providing structural information that goes beyond purely statistical descriptors and helping meta-models infer which algorithms are likely to perform well. Owing to their interpretability and strong predictive value, landmarkers have become increasingly prominent in recent AutoClustering frameworks~\citep{pimentel2018statistical, liu2021autocluster, elshawi2022tpe, treder2023ml2dac}, now ranking as the second most commonly used meta-feature family after statistical descriptors.

\paragraph{Model-based.} They are derived from models trained on the dataset, capturing properties of those models in response to the base datasets. In supervised settings, this typically includes indicators like tree depth, number of rules, or number of support vectors~\citep{vanschoren2018meta}. In AutoClustering, model-based meta-features can reflect clustering structure, geometric properties, or algorithmic behaviour beyond simple partitioning outcomes.

For instance, the MARCO-GE framework~\citep{COHENSHAPIRA2021824} applies this idea by encoding characteristics of a graph convolutional neural network (GCNN) trained on the dataset, using the learned topology and connectivity patterns as descriptors of the underlying base data manifold. Similarly, the AutoCluster system~\citep{liu2021autocluster} extracts model-based features by applying three different clustering algorithms: KMeans, Agglomerative, and OPTICS. Then, it records characteristics of their output, such as the compactness of KMeans clusters, the number of leaves in the hierarchical tree produced by Agglomerative clustering, and the reachability plot derived from OPTICS. These features serve to encapsulate structural complexity and clustering behaviour, enabling the meta-model to better understand which algorithms could align with the a new data's structure.

\paragraph{Complexity.} This family of meta-features captures the intrinsic difficulty of clustering a dataset. They include measures such as feature space overlap, Fisher discriminant ratio, density discontinuities, and boundary ambiguity~\citep{ho2002, lorena2019complex, vanschoren2018meta}. These descriptors quantify properties like how well-separated the data instances are or how clearly defined the cluster boundaries appear. In the context of AutoClustering, this family remains somewhat under explored because they are generally more computationally expensive than simple or statistical meta-features. Nevertheless, it has been used in exploratory and competitive systems. For instance, ~\citep{adam2015dealing} introduce the Constraint-Based Overlapping (CBO) value, designed to quantify ambiguity between clusters and serving as a proxy for dataset complexity. Other systems that incorporate complexity meta-features include~\citet{pimentel2019unsupervised}, \citet{treder2023ml2dac}, and~\citet{silva2024arxiv}.

Overall, these families provide a structured way to reason about and compare meta-feature sets across AutoClustering frameworks. As we will see in the next subsection, different systems emphasize different families based on their design goals and target domains.

\subsection{Meta-Feature Usage Across AutoClustering Frameworks}

\begin{figure}[!htb]
    \centering
    \includegraphics[width=1.05\linewidth]{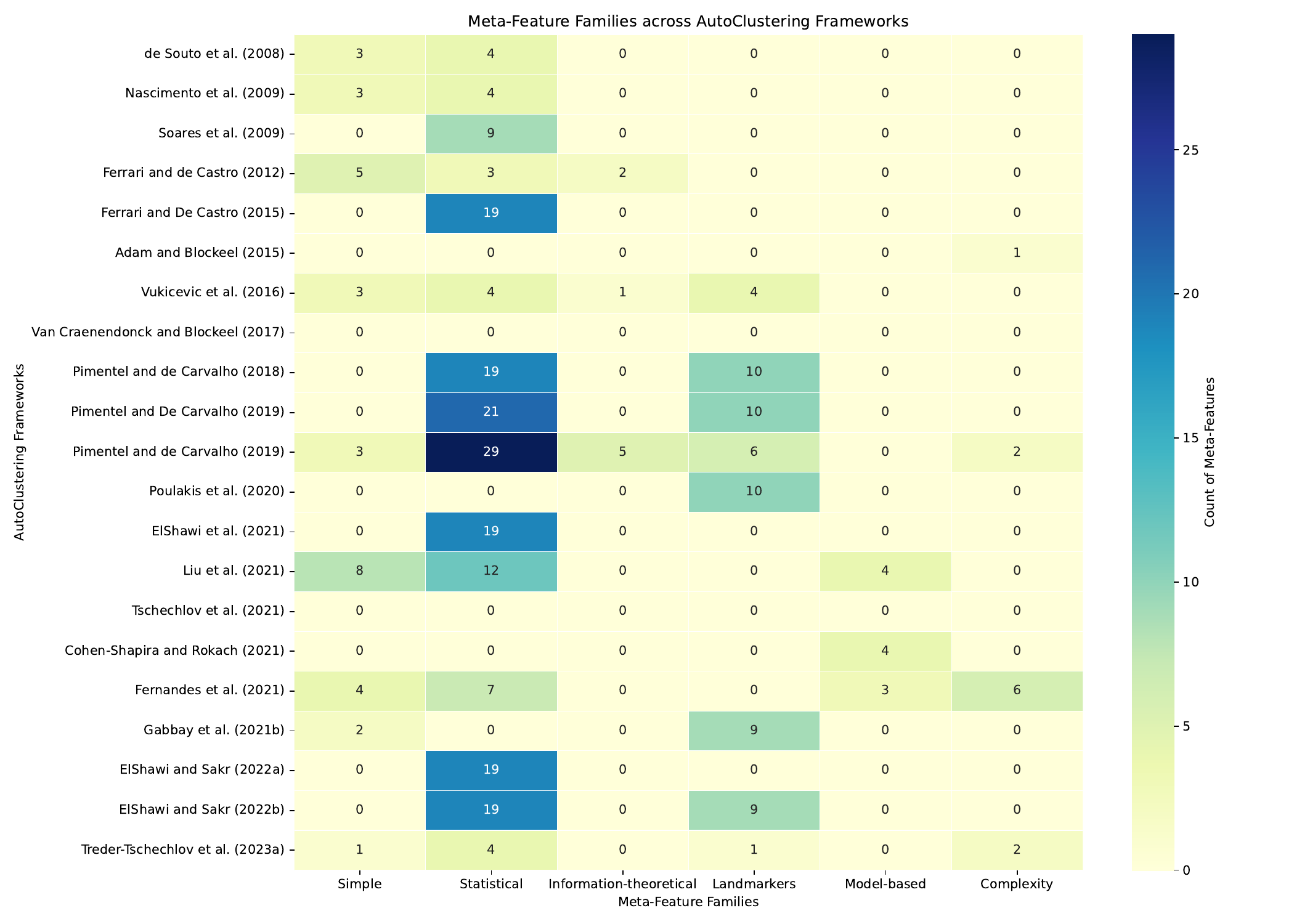}
    \caption{Heatmap showing the distribution of meta-feature families across various AutoClustering frameworks. The color intensity corresponds to the number of meta-features in each family for each framework. The frameworks are sorted ascending by publication date.}
    \label{fig:meta_feature_heatmap}
\end{figure}

Figure~\ref{fig:meta_feature_heatmap} summarizes the presence of each meta-feature family across the AutoClustering frameworks. Each cell indicates the number of features used per category.

The early studies, such as ~\citet{desouto2008ranking} and \citet{nascimento2009mining}, primarily focused on biological datasets characterized by high dimensionality, motivating the use of simple and statistical meta-features to capture distributional nuances. Subsequent works gradually incorporated landmarkers and more sophisticated information-theoretical measures to better characterize dataset structure. UCI datasets and other common benchmarks became widely adopted from around 2015 onward, facilitating broader system comparisons. Recent frameworks increasingly combine multiple meta-feature families, reflecting a trend towards richer meta-descriptions aimed at improving clustering pipeline selection and interpretability.

An analysis of the meta-feature usage across AutoClustering frameworks reveals several consistent trends. First, statistical meta-features are by far the most widely adopted across systems. Their general-purpose applicability, low computational overhead, and ease of extraction make them a default choice for describing dataset distributions and structures. Similarly, landmarking features have grown in popularity, particularly in frameworks developed after 2018. These features offer a practical means of capturing structural signals via the performance of simple clustering algorithms and are attractive due to their low cost and compatibility with unsupervised settings.

Model-based and complexity-related meta-features are comparatively rare, likely due to higher computation costs and limited support in standard meta-learning toolkits. Still, they may capture complementary signals about dataset geometry and clustering difficulty, making them promising targets for future work.

Simple and information-theoretic features appear more often, typically within larger sets, providing compact and interpretable summaries but offering limited discriminative power when used in isolation.

\section{Explainability of Meta-Models}
\label{sec:xai_mtl}

This section analyses how AutoClustering meta-models use dataset meta-features to produce pipeline decisions. 
Since recommendations are made in the \emph{meta-feature space} (datasets as instances), we adopt two complementary explanation scopes: 
a \emph{global} view to characterise the dominant decision logic learned across datasets, and a \emph{local} view to justify and inspect individual recommendations. 
Accordingly, we use DPG to reveal rule-like global structure and interactions, and SHAP to quantify instance-level feature attributions.

\subsection{Global Explanation} \label{sec:global_explainability}

The global analysis addresses two questions: 
(i) which meta-feature families consistently drive meta-model behaviour, and 
(ii) how stable these reliance patterns are across frameworks with different design choices (meta-model type, dataset sources, and meta-feature sets). 
To support cross-framework comparison, we summarise global importance in terms of DPG predicates ranked by LRC, and we report both the most influential and least influential predicates to expose concentration and redundancy effects.

\subsubsection{Framework Selection and Experimental Setup}

To analyse the global explainability of AutoClustering systems, we selected four representative frameworks: AutoClust~\citep{poulakis2020autoclust}, AutoCluster~\citep{liu2021autocluster}, ML2DAC~\citep{tschechlov2023ml2dac}, and PoAC~\citep{silva2024arxiv}. These frameworks were chosen based on two key criteria: (1) their reliance on meta-modeling, as opposed to meta-embedding approaches (see Section~\ref{sec:autoclustering_background} for a discussion of these paradigms), and (2) their reproducibility, specifically, whether they provided access to the meta-feature space and trained meta-models or at least the data required to reconstruct them.

While several frameworks described in the previous section use meta-learning, many did not offer sufficient information to reconstruct the meta-space or presented reproducibility issues. This limited their inclusion in our experiments.

\subsubsection{Discriminative Power of Meta-Features}
To assess the discriminative power of meta-features in AutoClustering, we analysed the ten most and least central predicates of the meta-model for each framework, ranked according to their LRC as computed by the DPG.

\paragraph{AutoClust.}
By design, AutoClust relies exclusively on a compact set of landmarker-based meta-features, consisting solely of internal Cluster Validity Indices (CVIs), including \textit{SIL}, \textit{DBS}, \textit{ch}, \textit{dunn}, \textit{cop}, and related measures. Despite the limited size of this feature set, a clear differentiation in discriminative power emerges when predicates are ranked by LRC.

A small subset of CVIs—most notably \textit{SIL}, \textit{DBS}, and \textit{ch}—consistently occupies the top-ranked positions, indicating a strong influence on the decision paths extracted from the DPGs. Conversely, other indices such as \textit{dunn}, \textit{cop}, and the \textit{c-index} predominantly appear among the bottom-ranked predicates, suggesting a comparatively minor contribution to the framework’s inferences.

Although the LRC distribution is relatively balanced due to the homogeneous nature of the feature family, the consistent dominance of only a few CVIs indicates redundancy within the set. This observation suggests that even in tightly constrained, landmarker-based systems, not all theoretically well-established metrics provide meaningful discriminative power in practice, and that selective pruning could improve both interpretability and computational efficiency.

\paragraph{AutoCluster.}
In contrast, \textbf{AutoCluster} primarily relies on \textit{statistical} and \textit{simple} dataset descriptors, such as \texttt{mean.sd}, \texttt{iq\_range.sd}, and \texttt{nr\_num}. These low-cost meta-features consistently appear among the most influential predicates, highlighting their strong discriminative capacity despite their simplicity. Although \textit{model-based} meta-features form a central part of AutoCluster’s original design, they are largely absent from the top-ranked predicates, with the notable exception of \texttt{reachability $>$ 7.51}.

This threshold indicates that datasets exceeding a moderate reachability value carry discriminative information relevant to AutoCluster’s decision-making. By comparison, other model-based predicates, such as \texttt{reachability $\leq$ 52.03} and \texttt{n\_leaves $>$ 3000}, appear among the bottom-ranked predicates. These extreme or highly dataset-specific thresholds likely limit their usefulness for general discrimination, explaining their low LRC values.

The bottom-ranked predicates are again dominated by statistical descriptors, indicating that even within a favoured feature family, discriminative power is unevenly distributed. Overall, these findings suggest that AutoCluster’s meta-model effectively concentrates on a small subset of stable, inexpensive features, while more complex descriptors contribute marginally, raising questions regarding the cost-effectiveness of their inclusion.

\paragraph{PoAC and ML2DAC.}
Conversely to both AutoClust and AutoCluster, \textbf{PoAC} adopts a more balanced design, incorporating meta-features from all five major families. However, despite the explicit inclusion of \textit{complexity-based} meta-features, none appear among the top-ranked predicates. A similar pattern is observed for ML2DAC, which includes two complexity-based features that likewise fail to achieve high LRC scores.

The consistent absence of complexity-based features from the most influential predicates suggests that, despite their theoretical appeal, such features may be either too computationally expensive or too unstable across datasets to provide robust discriminative value. This observation motivates a closer examination of their practical utility. We investigate this issue further in Section~\ref{sec:meta-model-ablation}, where ablated versions of PoAC’s meta-model are evaluated after removing time-intensive meta-features, allowing us to assess the resulting impact on predictive performance.

\begin{figure}[!htb]
\centering
\begin{subfigure}[t]{\textwidth}
    \centering
    \includegraphics[width=\textwidth]{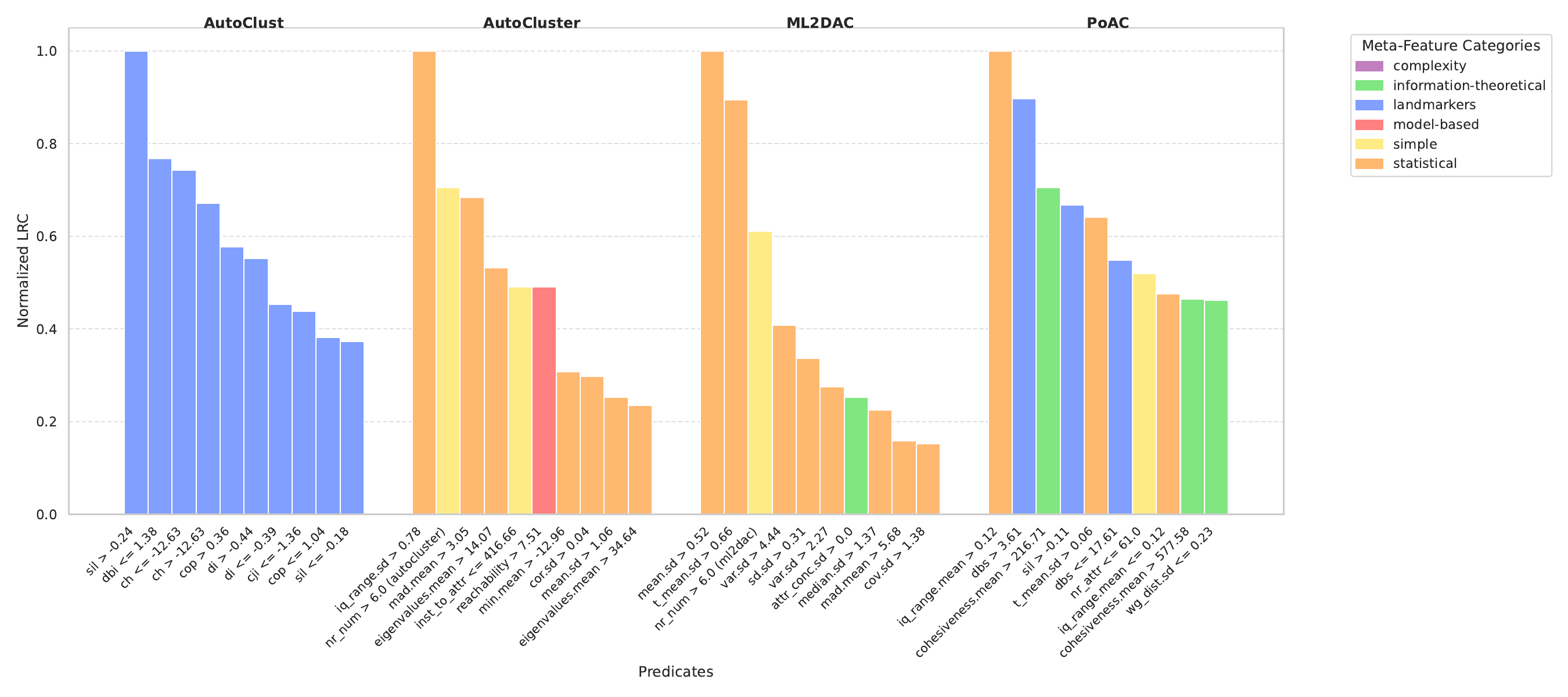}
    \caption{Top 10 most relevant meta-feature predicates per AutoClustering framework based on LRC.}
    \label{fig:autoclustering_lrc_top}
\end{subfigure}

\vspace{1em}

\begin{subfigure}[t]{\textwidth}
    \centering
    \includegraphics[width=\textwidth]{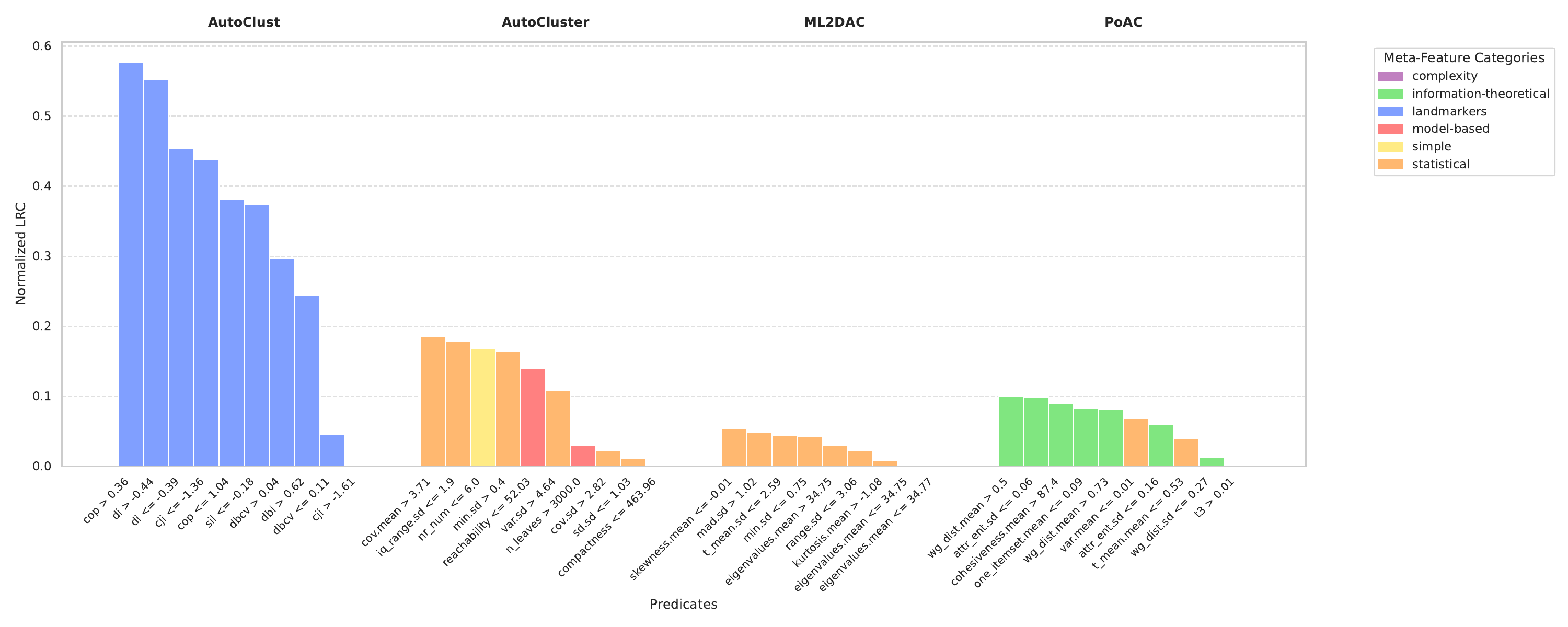}
    \caption{The 10 least relevant meta-feature predicates per framework, based on LRC.}
    \label{fig:autoclustering_lrc_bottom}
\end{subfigure}

\caption{Comparison of top and bottom meta-feature predicates per AutoClustering framework (\textit{AutoClust}, \textit{AutoCluster}, \textit{ML2DAC}, \textit{PoAC}) based on LRC. Top predicates (a) highlight features most embedded in the information flow, while bottom predicates (b) reflect less influential or potentially redundant aspects. Bars are color-coded by meta-feature category.}
\label{fig:autoclustering_lrc_combined}
\end{figure}

A complementary view is provided in the Appendix ~\ref{tab:grouped_predicates_sideways}, which groups the top predicates by meta-feature and indicates the degree of overlap across frameworks. Several features, such as \texttt{SIL}, \texttt{dbi}, \texttt{t\_mean.sd}, and \texttt{iq\_range}, appear in multiple frameworks, often with different threshold splits. This recurrence signals that certain descriptors (especially those related to internal structure and data dispersion) are broadly informative, regardless of the framework’s modeling paradigm.

Overall, while AutoClustering frameworks tend to adopt large and diverse meta-feature sets, only a narrow subset tends to drive model behaviour. The consistent underuse of certain families (notably complexity and model-based features), as well as the repeated reliance on a few general-purpose descriptors, motivates future work on meta-feature selection, redundancy reduction, and the development of more adaptive feature design strategies guided by model explainability.

\subsection{Local Explainability of Meta-Model Decisions}

While global explanations capture dominant patterns, AutoClustering systems are ultimately used on individual datasets. 
We therefore complement the global analysis with local explanations to answer: \emph{why did the meta-model produce this recommendation for this dataset?} 
Using SHAP, we quantify the contribution of each meta-feature value to a specific prediction in the meta-feature space. 
In addition to supporting case-by-case justification, this local view helps identify regime shifts where the model relies on different signals under different expected performance levels, and it can reveal whether global prominence translates into local impact.

\begin{figure}[!htb]
    \centering
    \includegraphics[width=0.85\textwidth]{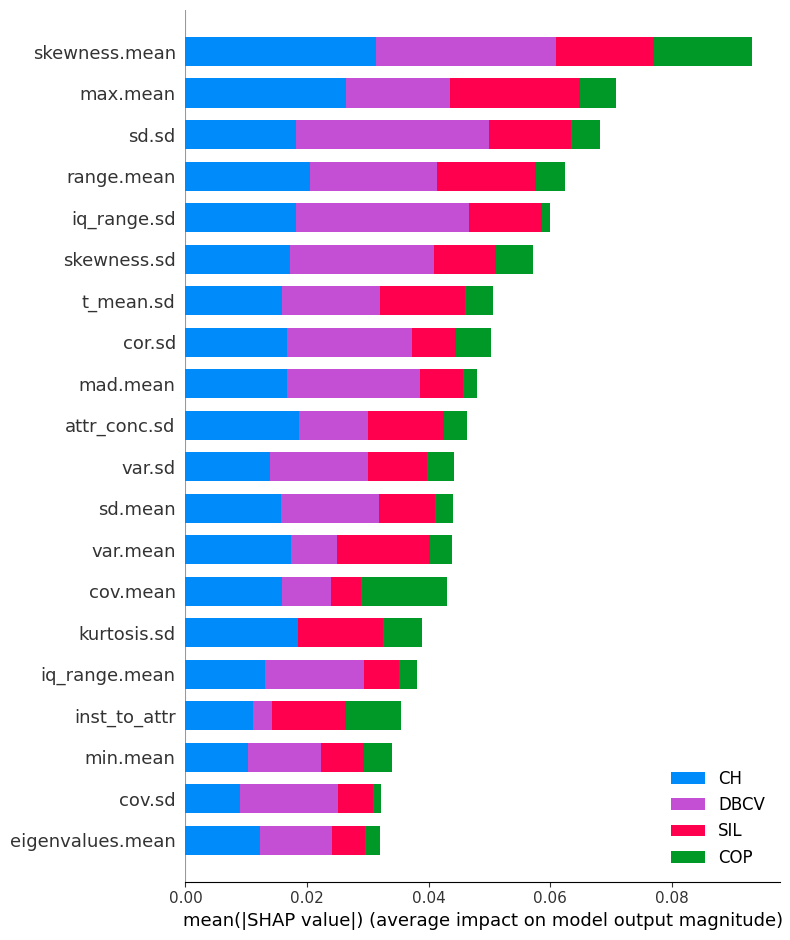}
    \caption{SHAP summary plot for ML2DAC showing the average contribution of each meta-feature to the model’s CVI recommendation. Colors denote the predicted CVI (CH, DBCV, SIL, COP).}
    \label{fig:ml2dac_shap_summary}
\end{figure}

\paragraph{ML2DAC: Explaining Meta-Model Preferences for CVIs.}  
Since ML2DAC casts the meta-modeling task as a classification problem, where the predicted classes correspond to different internal validation criteria (CH, DBCV, SIL and COP), we use SHAP to quantify how each meta-feature contributes to these CVI recommendations. As shown in Figure~\ref{fig:ml2dac_shap_summary}, DBCV and CH appear as the most frequently predicted CVIs, dominating the SHAP impact across top-ranked features. This suggests that ML2DAC implicitly favors these indices, possibly due to their consistent signal across datasets or stronger correlations with certain statistical descriptors. Moreover, we observe distinct feature-to-class affinities: for instance, \texttt{mad.mean} and \texttt{iq\_range.sd} are predominantly linked to DBCV recommendations, while \texttt{var.sd} and \texttt{cor.sd} tend to drive CH predictions. Conversely, COP surfaces far less frequently, and only a handful of features, such as \texttt{skewness.mean}, exhibit meaningful association with it. These patterns indicate that ML2DAC’s classification meta-model not only relies on a compact set of high-impact statistical features, but also aligns those selectively with the characteristics emphasized by individual CVIs.

\paragraph{PoAC} employs regression to predict clustering quality (ARI) based on an ensemble of meta-features and internal validation metrics. To understand which meta-features PoAC’s model \( f(x) \) prioritizes when assigning high ARI scores, we performed a cohort-level SHAP analysis on its recommended pipelines for 25 UCI datasets (sourced from~\cite{tpot-clustering}).

\begin{figure}[!htb]
    \centering
    \includegraphics[width=0.6\textwidth]{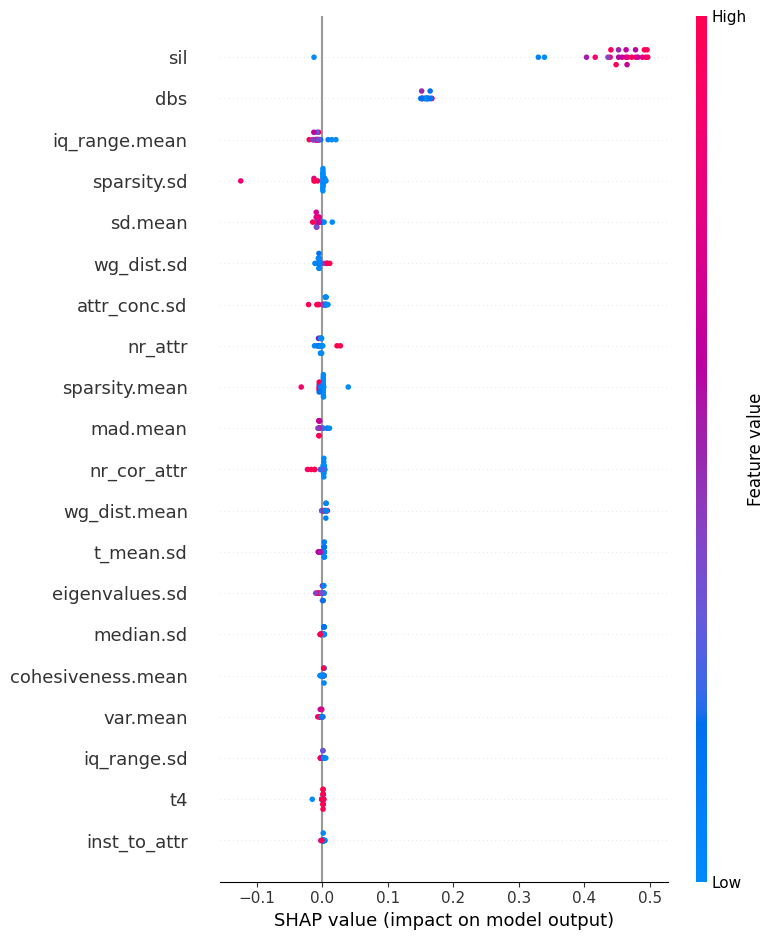}
    \caption{Cohort-level SHAP summary plot analysing PoAC's recommended clustering pipelines across 25 datasets. Each point represents a dataset/pipeline meta-feature value (color: red=high, blue=low) and its impact on the prediction (x-axis: SHAP value). Meta-features are ranked by mean absolute SHAP value.}
    \label{fig:shap_cohort}
\end{figure}

Figure~\ref{fig:shap_cohort} shows that PoAC’s predictions are primarily influenced by two internal validation metrics, \textit{SIL} and \textit{DBS}, both of which exhibit large SHAP magnitudes (up to \(+0.5\)). This indicates that the model relies strongly on internal consistency signals to infer clustering quality, even though these metrics are not explicitly part of its supervised training objective.

While Figure~\ref{fig:shap_cohort} does not provide causal explanations, they do reveal clear trends. Higher values of \textit{SIL} are generally associated with increased predicted ARI, whereas higher values of \textit{DBS} tend to decrease it. This aligns with standard interpretations in clustering evaluation, where well-separated clusters (high \textit{SIL}) are preferred and scattered or overlapping clusters (high \textit{DBS}) are penalized. The model appears to have internalized these domain heuristics during training.

Other meta-features, including statistical descriptors such as \textit{nr\_attr} and \textit{wg\_dist.mean}, show negligible impact on predictions (with absolute SHAP values below 0.1). This indicates that PoAC's regression model concentrates its decision boundary around a narrow subset of influential signals, potentially limiting its sensitivity to broader dataset characteristics.

\begin{figure}[ht]
    \centering
    \begin{subfigure}[t]{0.45\textwidth}
        \centering
        \includegraphics[width=\textwidth]{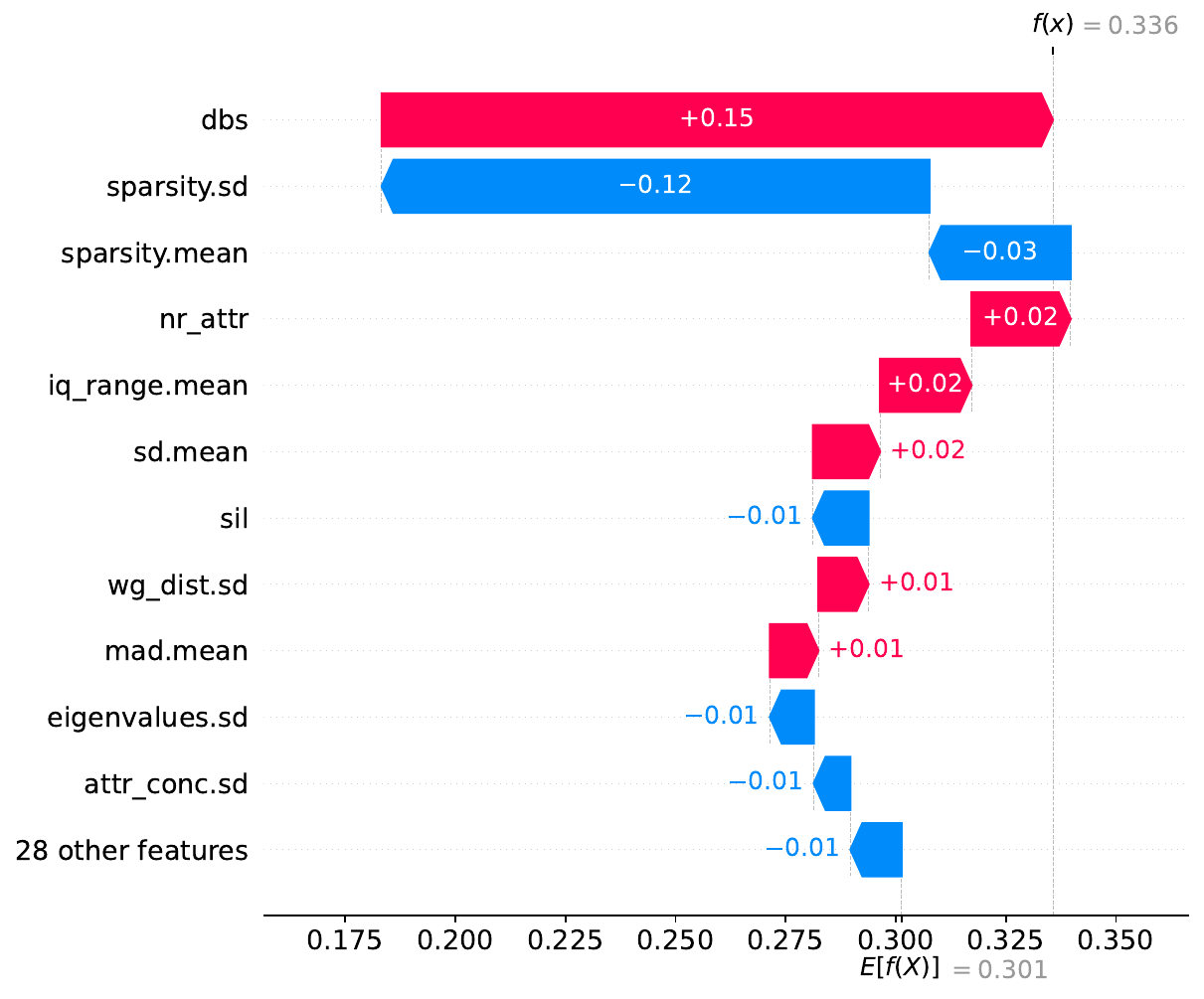}
        \caption{Low predicted ARI pipeline (\(f(x) = 0.336\)) for \textit{thy}.}
        \label{fig:thy_bad}
    \end{subfigure}
    \hfill
    \begin{subfigure}[t]{0.45\textwidth}
        \centering
        \includegraphics[width=\textwidth]{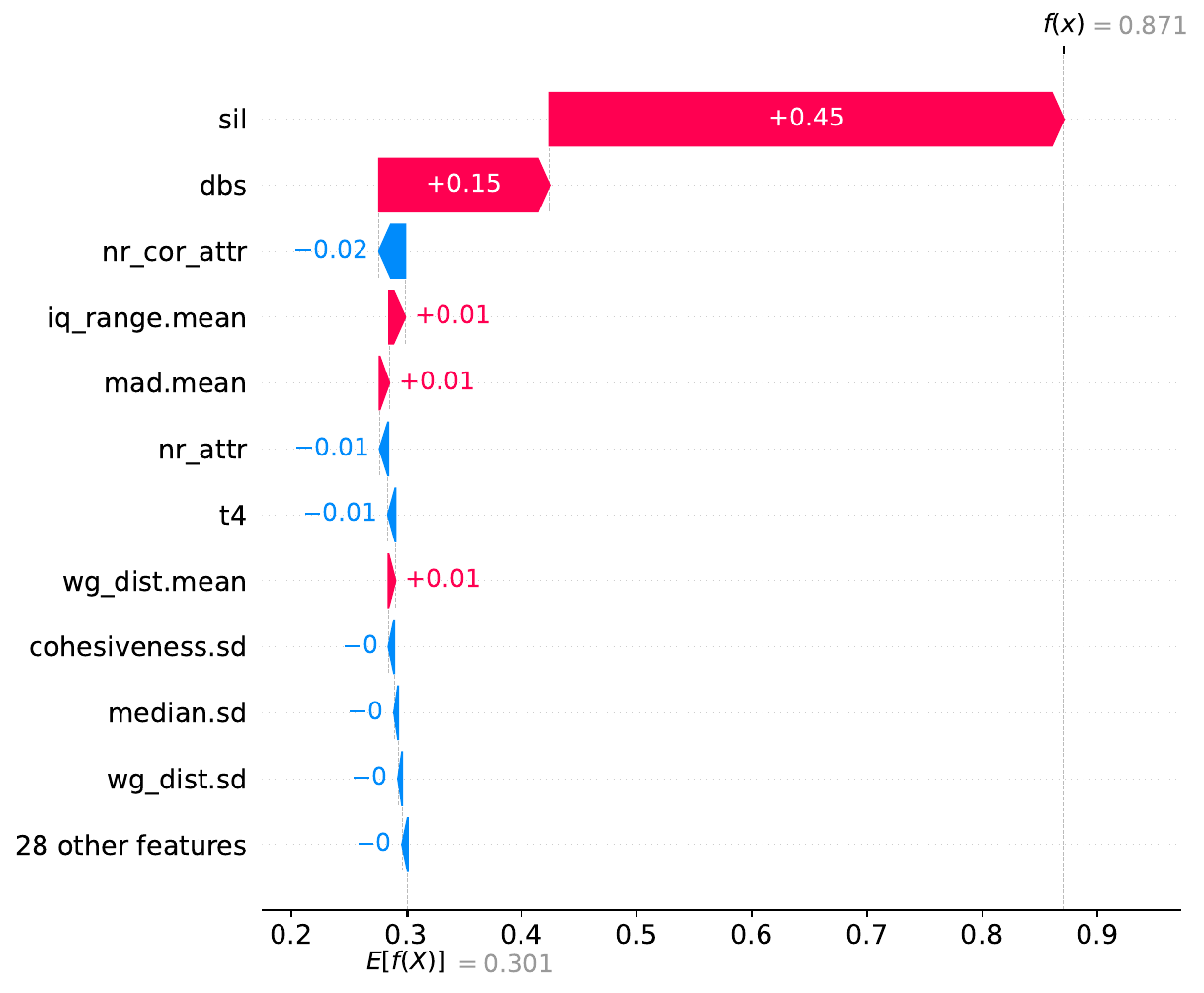}
        \caption{High predicted ARI pipeline (\(f(x) = 0.871\)) for \textit{thy}.}
        \label{fig:thy_good}
    \end{subfigure}
    \caption{SHAP explanations for two PoAC-predicted pipelines on the \textit{thy} dataset, representing low and high clustering quality.}
    \label{fig:local_shap}
\end{figure}

To further understand PoAC’s shifting reliance on meta-features across performance regimes, we compared SHAP explanations for two pipelines applied to the thy dataset—one with high predicted ARI (\( f(x) \)=0.871) and one with low (\( f(x) \)=0.336). As seen in Figure~\ref{fig:local_shap}, the high-performing pipeline strongly leverages internal validation metrics: \textit{SIL} contributes \(+0.45\), followed by modest positive influence from features like \textit{DBS}, \textit{mad.mean}, and \textit{wg\_dist.mean}. This confirms the model’s dependence on internal consistency signals when assigning high ARI scores. By comparison, the low-scoring pipeline shows weakened influence from \textit{SIL} and \textit{DBS}, with \textit{sparsity.sd} and \textit{sparsity.mean} emerging as context-specific signals. Their small but nonzero contributions reflect a pivot in PoAC’s inference strategy: in the absence of strong internal validation cues, dataset-inherent properties such as attribute sparsity take on explanatory weight. This contrast illustrates how PoAC dynamically adjusts its feature importance profile based on the expected clustering quality.

\section{Meta-model Ablation}
\label{sec:meta-model-ablation}

This section evaluates whether the explanation evidence reported in Section~\ref{sec:xai_mtl} can be used to make AutoClustering systems more efficient. 
In practice, meta-feature extraction is part of the inference pipeline: it must be executed whenever a new dataset arrives, and it can dominate end-to-end latency in time-sensitive or large-scale settings~\citep{brazdil2022metalearning, reif2012meta}. 
At the same time, our global analysis indicates that meta-model decisions are often driven by a small subset of descriptors, suggesting that some features are redundant or weakly used. 
We therefore perform a controlled ablation study to quantify the trade-off between meta-feature extraction cost and predictive performance, and to assess whether explainability-guided pruning yields practical gains without materially degrading accuracy.

\subsection{Ablation Setup}
We focus on PoAC~\citep{silva2024arxiv} because it provides a complete and reproducible meta-feature space and meta-model training pipeline. 
We evaluate three versions of the PoAC meta-feature set on the same 6,000 synthetic datasets released with the framework~\citep{silva2024arxiv}. 
The three sets are:
(i) \emph{PoAC Full}, the original set of 39 meta-features; 
(ii) \emph{PoAC DPG}, a reduced set of 24 meta-features obtained by filtering descriptors whose associated predicates have negligible LRC; and 
(iii) \emph{PoAC Core}, a compact set composed of the 10 most central meta-features according to average LRC. 
This design isolates the effect of feature selection while keeping the learning algorithm and evaluation protocol unchanged.

\begin{figure}[!htb]
    \centering
    \includegraphics[width=1\textwidth]{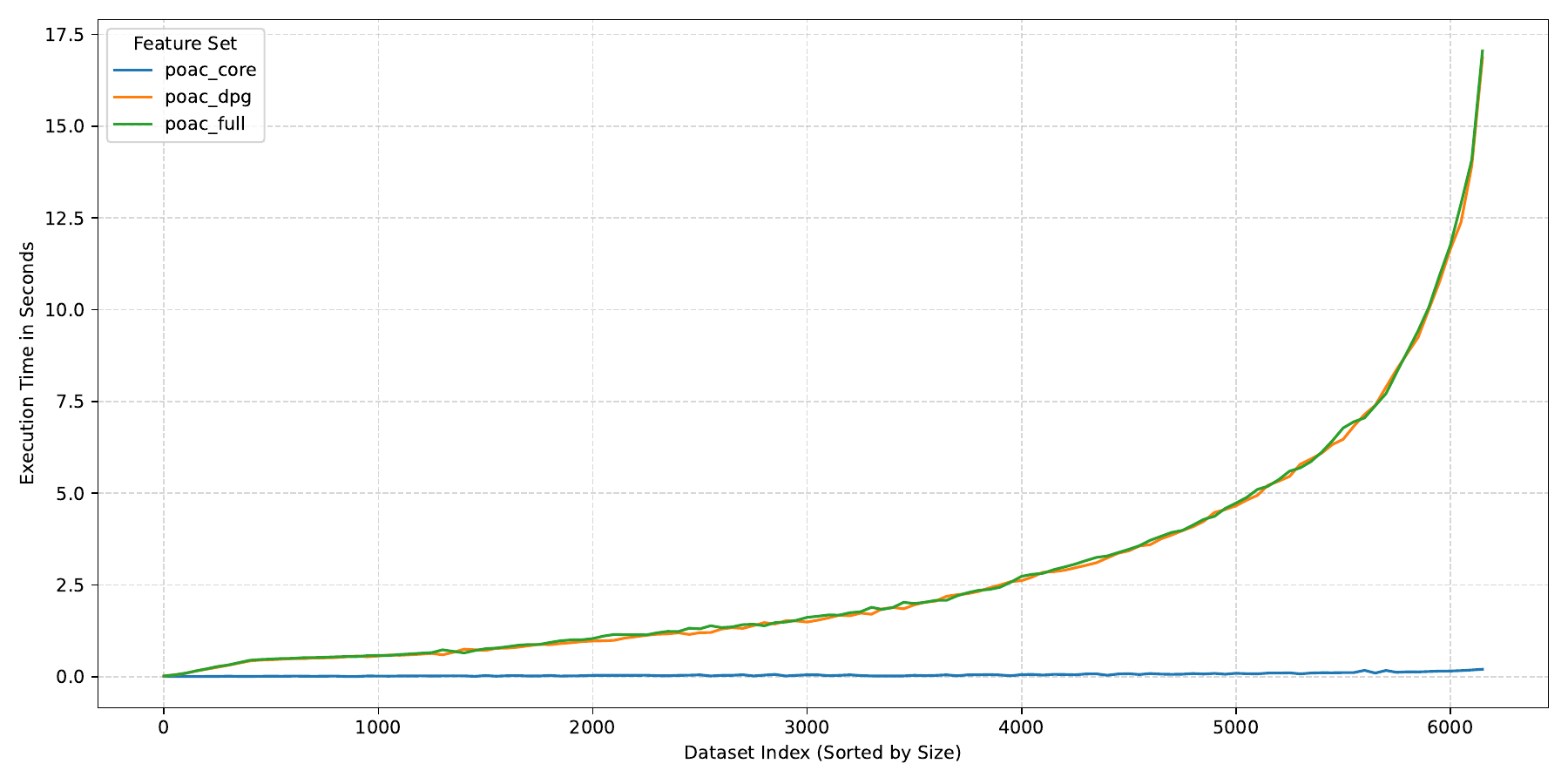}
    \caption{
        Ablation experiment comparing the meta-feature extraction time across three different feature sets: 
        (i) \texttt{poac\_full} the complete set of 39 meta-features used by Poac; 
        (ii) \texttt{poac\_dpg} a reduced set of 24 meta-features with a minimum LRC value; 
        (iii) \texttt{poac\_core} a minimal set of the 10 most central meta-features based on LRC.
    }
    \label{fig:poac_ablation}
\end{figure}

Figure~\ref{fig:poac_ablation} reports the total extraction time for each set across the 6,000 datasets. 
PoAC Full requires approximately 4 h 54 m, while PoAC DPG remains comparable at 4 h 49 m. 
By contrast, PoAC Core reduces extraction time to under 7 minutes, illustrating that a small subset of descriptors accounts for most of the computational cost.

\subsection{Performance Impact}

To measure the effect on prediction quality, we train a Random Forest regressor on each feature subset and evaluate it with 10-fold cross-validation on the 100 synthetic test datasets provided by~\citep{silva2024arxiv}. 
Table~\ref{tab:poac_results} summarises the results. 
As expected, PoAC Full provides the best overall performance. 
However, both reduced sets remain competitive: PoAC DPG increases RMSE by 6.3\% and MAE by 8.3\% relative to the full set, while PoAC Core yields an RMSE increase of 12\% and still achieves $R^2>0.82$. 
These results indicate that most of the predictive signal is retained after substantial feature reduction, especially when the retained features are chosen based on global relevance.

\begin{table}[!htb]
\centering
\begin{tabular}{lccccc}
\hline
\textbf{Meta-Feature Set} & \textbf{RMSE} ↓ & \textbf{MAE} ↓ & \textbf{R\textsuperscript{2}} ↑ & \textbf{Extraction Time} \\
\hline
PoAC Full  & \textbf{0.0735} & \textbf{0.0521} & \textbf{0.8644} & 4h 54m \\
PoAC DPG   & 0.0782 & 0.0564 & 0.8465 & 4h 49m \\
PoAC CORE  & 0.0825 & 0.0600 & 0.8291 & \textbf{0h 7m} \\
\hline
\end{tabular}
\caption{
Cross-validated performance of Random Forest regressors using different feature subsets.
Lower values are better for RMSE and MAE; higher values are better for R\textsuperscript{2}.
The last column shows the total time required to extract each meta-feature set across all 6,000 datasets.
}
\label{tab:poac_results}
\end{table}

The ablation study shows that explainability can be used as an engineering signal rather than a purely descriptive tool. 
Selecting features based on global relevance yields large reductions in extraction time while preserving most of the meta-model’s predictive accuracy, which is directly relevant for scaling meta-learning pipelines and for deployment in compute-constrained environments.

\section{Discussion}
\label{sec:discussion}

This work addressed a practical gap in AutoClustering: understanding how dataset meta-features drive the recommendations produced by meta-models, and what this implies for transparency, reliability, and efficiency in unsupervised AutoML. While much prior work has focused on proposing new meta-features or improving predictive performance, the decision logic that links meta-features to recommended pipelines is rarely examined. By combining a unified meta-feature taxonomy with complementary global and local analyses, we provide a structured view of how current AutoClustering meta-models behave in practice. 

Across the analysed frameworks, three consistent observations emerge: (i) decision-making is concentrated in a small subset of descriptors despite large feature sets, (ii) many recommendations are strongly shaped by internal-validation signals and related systematic preference effects, and (iii) explanation evidence can be translated into concrete efficiency gains through feature-set reduction.

Our global analysis using Decision Predicate Graphs reveals a striking concentration of influence across all examined frameworks. Despite the widespread adoption of large and diverse meta-feature sets, only a narrow subset consistently drives meta-model decisions. In particular, statistical descriptors and landmarker-based features dominate decision paths, while complexity-based and model-based meta-features are rarely influential. This pattern holds across frameworks with distinct design philosophies, suggesting that redundancy is not an isolated implementation artifact but a structural characteristic of current AutoClustering meta-models. These findings challenge the assumption that increasing meta-feature diversity necessarily improves generalization and instead point toward diminishing returns beyond a core set of stable descriptors.

Local explainability analyses using SHAP further refine this picture by exposing how meta-features contribute to individual recommendations. For ML2DAC, we observe clear alignments between specific statistical descriptors and the preferred internal validation criteria, indicating that the meta-model learns feature-to-objective affinities rather than treating CVIs interchangeably. In PoAC, SHAP explanations reveal a strong dependence on internal validation signals, particularly SIL and DBS, when predicting clustering quality. While this behaviour aligns with established clustering heuristics, it also exposes an implicit bias toward internal consistency measures, potentially limiting sensitivity to broader dataset characteristics. Importantly, local explanations show that PoAC dynamically shifts its reliance on meta-features depending on the expected clustering quality, demonstrating that even compact decision logic can remain context-sensitive.

Building on these insights, the meta-model ablation study provides concrete evidence that explainability can be leveraged not only for interpretation but also for optimization. By pruning meta-features based on global relevance scores, we achieve dramatic reductions in extraction time with only marginal losses in predictive performance. The most compact configuration, using fewer than one third of the original meta-features, preserves over 95 percent of the predictive accuracy while reducing extraction time by orders of magnitude. This result underscores a key practical implication of our analysis: explainability-driven ablation offers a principled pathway toward scalable and resource-efficient AutoML systems. Table \ref{tab:takeaway} present a summary of our findings. 

\begin{table}[!htb]
\centering
\caption{Summary of findings for AutoClustering design.}
\label{tab:takeaway}
\small
\setlength{\tabcolsep}{5pt}
\begin{tabular}{p{4.5cm} p{10cm}}
\toprule
\textbf{Observation} & \textbf{Why it matters} \\
\midrule
Decision influence is concentrated in a small subset of meta-features (DPG). 
& Large meta-feature sets often introduce redundancy and do not necessarily improve generalisation beyond a stable core set. \\
\addlinespace
Internal-validation signals strongly shape recommendations (SHAP), indicating systematic preference effects. 
& Recommendations may inherit assumptions encoded by CVIs and landmarkers, which can diverge from user intent or domain semantics in unsupervised settings. \\
\addlinespace
Global and local explanations provide complementary evidence (DPG vs.\ SHAP). 
& Global structure can highlight dominant decision logic, while local attributions justify individual recommendations and reveal regime-dependent feature reliance. \\
\addlinespace
Explainability-guided pruning yields large efficiency gains with modest accuracy loss (ablation). 
& Meta-feature extraction can dominate end-to-end runtime; pruning based on relevance provides a principled route to faster, more deployable systems. \\
\bottomrule
\end{tabular}
\end{table}

Taken together, the findings suggest several design implications for future AutoClustering systems. First, explanation should be considered a design-time requirement, not only a post-hoc diagnostic: exposing dominant decision structures helps debugging and supports auditing of systematic preference effects. Second, meta-feature engineering should be guided by measured contribution and cost: expensive descriptors should be included only when they provide demonstrable marginal benefit over the core set. Third, since internal-validation signals frequently dominate decisions, future work should study how meta-models balance CVI-based cues with broader dataset descriptors, especially when user intent or domain semantics differ from the geometric assumptions encoded by many CVIs.

This study also has limitations. Our experiments are restricted to frameworks whose meta-feature spaces and trained meta-models could be reliably reconstructed, which reduces coverage of recent embedding-based approaches. In addition, the ablation study is conducted on the PoAC synthetic benchmark suite; although it enables controlled comparisons, transfer to fully real-world meta-datasets should be further validated. Finally, while DPG and SHAP provide complementary evidence, extending global structural analysis to neural or representation-learning meta-models remains an open direction. Addressing these points would broaden the applicability of the proposed analysis and strengthen guidance for next-generation AutoClustering systems.

\section{Conclusions}
\label{sec:conclusion}
This paper examined AutoClustering from the perspective of the meta-learning layer, asking how dataset meta-features actually drive the recommendations produced by clustering meta-models. We contributed (i) a structured taxonomy covering meta-features and dataset sources across 22 AutoClustering frameworks, (ii) a two-level explainability analysis that combines global decision structure extraction via Decision Predicate Graphs with local attributions via SHAP, and (iii) an explainability-guided ablation study showing that substantial reductions in meta-feature extraction cost can be achieved with only modest losses in predictive accuracy. 

Empirically, our results indicate that current AutoClustering meta-models concentrate their decisions on a small subset of largely statistical and landmarker-based descriptors, while more costly feature families are often weakly utilised. These findings suggest that explainability can serve as an engineering signal: it supports auditing of systematic preference effects, motivates principled meta-feature pruning, and helps move AutoClustering from opaque recommendation to inspectable decision-making. Future work should extend these analyses to embedding-based and neural meta-learning approaches and validate the proposed efficiency--accuracy trade-offs on broader real-world meta-datasets.

\bibliographystyle{unsrtnat}
\bibliography{references} 

\newpage
\appendix
\section{Appendix}
\subsection{Meta-feature family coverage across frameworks}
\label{app:mf_usage_discussion}

Table~\ref{tab:meta_feature_usage} summarises the meta-feature families reported by the 22 AutoClustering frameworks reviewed in this study, grouped according to the taxonomy introduced in Section~\ref{sec:taxonomy}. 
Each entry corresponds to the number of meta-features in a given family explicitly used by the framework. 
The table is intended to support two purposes: (i) to provide a compact, reproducible record of how meta-feature families are instantiated in prior work, and (ii) to make the heterogeneity and redundancy of meta-feature design choices visible at a glance.

Our results in Sections~\ref{sec:global_explainability}--\ref{sec:meta-model-ablation} qualify this assumption by showing that, even when many families are computed, meta-model behaviour often concentrates on a comparatively small core of influential descriptors.

\begin{table*}[!htb]
\centering
\scriptsize
\caption{Meta-feature family usage across AutoClustering frameworks (2008–2024).}
\label{tab:meta_feature_usage}
\begin{tabular}{lcccccc}
\toprule
\textbf{Framework} & \rotatebox{90}{\textbf{Simple}} & \rotatebox{90}{\textbf{Statistical}} & \rotatebox{90}{\textbf{Info.-Theoretical}} & \rotatebox{90}{\textbf{Landmarkers}} & \rotatebox{90}{\textbf{Model-Based}} & \rotatebox{90}{\textbf{Complexity}} \\
\midrule
Souto (2008)           & 3 & 4  & 0 & 0  & 0 & 0 \\
Nascimento (2009)      & 3 & 4  & 0 & 0  & 0 & 0 \\
Soares (2009)          & 0 & 9  & 0 & 0  & 0 & 0 \\
Ferrari (2012)         & 5 & 3  & 2 & 0  & 0 & 0 \\
Ferrari (2015)         & 0 & 19 & 0 & 0  & 0 & 0 \\
CBOvalue (2015)        & 0 & 0  & 0 & 0  & 0 & 1 \\
Vukicevic (2016)       & 3 & 4  & 1 & 4  & 0 & 0 \\
COBS (2017)            & 0 & 0  & 0 & 0  & 0 & 0 \\
Pimentel (2018)        & 0 & 19 & 0 & 10 & 0 & 0 \\
Pimentel (2019a)       & 0 & 21 & 0 & 10 & 0 & 0 \\
Pimentel (2019b)       & 3 & 29 & 5 & 6  & 0 & 2 \\
Autoclust (2020)       & 0 & 0  & 0 & 10 & 0 & 0 \\
cSmartML (2021)        & 0 & 19 & 0 & 0  & 0 & 0 \\
AutoCluster (2021)     & 8 & 12 & 0 & 0  & 4 & 0 \\
AutoML4Clust (2021)    & 0 & 0  & 0 & 0  & 0 & 0 \\
Marco-GE (2021)        & 0 & 0  & 0 & 0  & 4 & 0 \\
Fernandes (2021)       & 0 & 0  & 0 & 0  & 0 & 0 \\
Gabbay (2021)          & 2 & 0  & 0 & 9  & 0 & 0 \\
cSmartML-GB (2022)     & 0 & 19 & 0 & 0  & 0 & 0 \\
TPE-Autoclust (2022)   & 0 & 19 & 0 & 9  & 0 & 0 \\
ML2DAC (2023)          & 1 & 4  & 0 & 1  & 0 & 2 \\
PoAC (2024)            & 4 & 10 & 6 & 3  & 0 & 3 \\
\bottomrule
\end{tabular}
\end{table*}

\subsection{Top DPG predicates and cross-framework comparability}
\label{app:top_predicates_lrc_discussion}

Table~\ref{tab:top_predicates_lrc} reports the ten most central predicates extracted by Decision Predicate Graphs (DPG) for each analysed framework, ranked by Local Reaching Centrality (LRC). 
Each predicate is a human-readable rule of the form \texttt{feature} \,$\circ$\, \texttt{threshold} that appears in the reconstructed global decision structure. 
We report both the original LRC values and a per-framework normalised version (max-normalised to 1.0) to support two complementary readings: 
(i) \emph{within-framework ranking}, where normalised LRC highlights which predicates dominate a given model, and 
(ii) \emph{cross-framework inspection}, where the predicate \emph{categories} and recurring feature names provide a qualitative notion of common drivers across different meta-models.

This table therefore complements Figure~\ref{fig:autoclustering_lrc_combined} by making the dominant decision rules explicit and by highlighting the degree to which current meta-models concentrate on a small number of stable, repeatedly used descriptors.
In the main text, these observations motivate the local analyses (Section~\ref{sec:xai_mtl}) and the explainability-guided ablation study (Section~\ref{sec:meta-model-ablation}), which evaluate whether the globally prominent signals also drive individual recommendations and whether weakly used descriptors can be removed with limited performance impact.

\begin{table}[htbp]
\centering
\tiny
\caption{Top meta-feature predicates per AutoClustering framework based on original and normalized LRC.}
\label{tab:top_predicates_lrc}
\begin{tabularx}{\textwidth}{l l r r l}
\toprule
\textbf{Framework} & \textbf{Predicate} & \textbf{LRC} & \textbf{LRC normalized} & \textbf{MF Category} \\
\midrule
autoclust   & \texttt{SIL > -0.24}                & 33.0543   & 1.000   & Landmarkers \\
autoclust   & \texttt{DBI <= 1.38}                & 30.2646   & 0.7678  & Landmarkers \\
autoclust   & \texttt{CH <= -12.63}               & 29.9616   & 0.7426  & Landmarkers \\
autoclust   & \texttt{CH > -12.63}                & 29.1075   & 0.6715  & Landmarkers \\
autoclust   & \texttt{COP > 0.36}                 & 27.9701   & 0.5768  & Landmarkers \\
autoclust   & \texttt{DI > -0.44}                 & 27.6708   & 0.5519  & Landmarkers \\
autoclust   & \texttt{DI <= -0.39}                & 26.4897   & 0.4536  & Landmarkers \\
autoclust   & \texttt{CJI <= -1.36}               & 26.2989   & 0.4377  & Landmarkers \\
autoclust   & \texttt{COP <= 1.04}                & 25.6228   & 0.3814  & Landmarkers \\
autoclust   & \texttt{SIL <= -0.18}               & 25.5206   & 0.3729  & Landmarkers \\
\midrule
autocluster & \texttt{iq\_range.sd > 0.78}        & 0.7413    & 1.000   & Statistical \\
autocluster & \texttt{nr\_num > 6.0}              & 0.5581    & 0.7046  & Simple \\
autocluster & \texttt{mad.mean > 3.05}            & 0.5454    & 0.6841  & Statistical \\
autocluster & \texttt{eigenvalues.mean > 14.07}   & 0.4513    & 0.5325  & Statistical \\
autocluster & \texttt{inst\_to\_attr <= 416.66}   & 0.4256    & 0.4910  & Simple \\
autocluster & \texttt{reachability > 7.51}        & 0.4255    & 0.4908  & Landmarkers \\
autocluster & \texttt{min.mean > -12.96}          & 0.3120    & 0.3080  & Statistical \\
autocluster & \texttt{cor.sd > 0.04}              & 0.3057    & 0.2978  & Statistical \\
autocluster & \texttt{mean.sd > 1.06}             & 0.2779    & 0.2530  & Statistical \\
autocluster & \texttt{eigenvalues.mean > 34.64}   & 0.2667    & 0.2349  & Statistical \\
\midrule
ml2dac      & \texttt{mean.sd > 0.52}             & 0.8927    & 1.000   & Statistical \\
ml2dac      & \texttt{t\_mean.sd > 0.66}          & 0.8179    & 0.8942  & Statistical \\
ml2dac      & \texttt{nr\_num > 6.0}              & 0.6177    & 0.6113  & Simple \\
ml2dac      & \texttt{var.sd > 4.44}              & 0.4743    & 0.4085  & Statistical \\
ml2dac      & \texttt{sd.sd > 0.31}               & 0.4238    & 0.3372  & Statistical \\
ml2dac      & \texttt{var.sd > 2.27}              & 0.3803    & 0.2756  & Statistical \\
ml2dac      & \texttt{attr\_conc.sd > 0.0}        & 0.3635    & 0.2519  & Information-theoretical \\
ml2dac      & \texttt{median.sd > 1.37}           & 0.3446    & 0.2252  & Statistical \\
ml2dac      & \texttt{mad.mean > 5.68}            & 0.2976    & 0.1588  & Statistical \\
ml2dac      & \texttt{cov.sd > 1.38}              & 0.2930    & 0.1522  & Statistical \\
\midrule
poac        & \texttt{iq\_range.mean > 0.12}      & 3.5425    & 1.000   & Statistical \\
poac        & \texttt{DBS > 3.61}                 & 3.4105    & 0.8972  & Landmarkers \\
poac        & \texttt{cohesiveness.mean > 216.71} & 3.1642    & 0.7055  & Information-theoretical \\
poac        & \texttt{SIL > -0.11}                & 3.1161    & 0.6681  & Landmarkers \\
poac        & \texttt{t\_mean.sd > 0.06}          & 3.0821    & 0.6416  & Statistical \\
poac        & \texttt{DBS <= 17.61}               & 2.9620    & 0.5481  & Landmarkers \\
poac        & \texttt{nr\_attr <= 61.0}           & 2.9261    & 0.5202  & Simple \\
poac        & \texttt{iq\_range.mean <= 0.12}     & 2.8692    & 0.4758  & Statistical \\
poac        & \texttt{cohesiveness.mean > 577.58} & 2.8539    & 0.4640  & Information-theoretical \\
poac        & \texttt{wg\_dist.sd <= 0.23}        & 2.8518    & 0.4623  & Information-theoretical \\
\bottomrule
\end{tabularx}
\end{table}

\subsection{AutoCluster meta-feature implementation details}
\label{app:autocluster_meta_features}

Table~\ref{app:autocluster_mfs} lists the 24 meta-features implemented in AutoCluster, grouped by category. 
We include this table for two reasons. 
First, AutoCluster is frequently referenced as a representative meta-learning approach for clustering recommendation, yet publications often describe its descriptors only at a high level. 
Second, our explainability results (Section~\ref{sec:global_explainability}) depend on how these features are instantiated, since DPG predicates and SHAP attributions operate directly on the computed meta-feature values.

\begin{table}[htbp]
\centering
\caption{The implemented meta-features in AutoCluster.}
\label{app:autocluster_mfs}
\begin{tabular}{llp{10cm}}
\toprule
\textbf{Category} & \textbf{No.} & \textbf{Meta-feature} \\
\midrule
\multirow{8}{*}{Simple} 
 & 1  & Number of instances \\
 & 2  & Log number of instances \\
 & 3  & Number of features \\
 & 4  & Log number of features \\
 & 5  & Dataset ratio \\
 & 6  & Log dataset ratio \\
 & 7  & Inverse dataset ratio \\
 & 8  & Inverse log dataset ratio \\
\midrule
\multirow{8}{*}{Statistical} 
 & 9  & Kurtosis max \\
 & 10 & Kurtosis min \\
 & 11 & Kurtosis mean \\
 & 12 & Kurtosis standard deviation \\
 & 13 & Skewness max \\
 & 14 & Skewness min \\
 & 15 & Skewness mean \\
 & 16 & Skewness standard deviation \\
\midrule
\multirow{3}{*}{PCA} 
 & 17 & PCA 95\% explained variance \\
 & 18 & Kurtosis of the first principal component \\
 & 19 & Skewness of the first principal component \\
\midrule
Data Distribution 
 & 20 & Hopkins statistic \\
\midrule
\multirow{4}{*}{Landmarker} 
 & 21 & Distance to closest cluster center (Kmeans) \\
 & 22 & Number of leaves (Agglomerative Clustering) \\
 & 23 & Maximum reachability distance (OPTICS) \\
 & 24 & Maximum core distance (OPTICS) \\
\bottomrule
\end{tabular}
\end{table}

\subsection{Cross-framework overlap of influential predicates}
\label{app:grouped_predicates_discussion}

Table~\ref{tab:grouped_predicates_sideways} aggregates the most relevant DPG predicates by meta-feature family and meta-feature name, and reports in which frameworks each predicate appears. 
This table complements Table~\ref{tab:top_predicates_lrc} by shifting the focus from per-framework rankings to \emph{cross-framework overlap}: rather than listing the top predicates separately for each meta-model, it highlights which meta-features recur as influential signals across different AutoClustering designs.

\begin{sidewaystable}[!htb]
\small
\setlength{\tabcolsep}{4pt}
\centering
\caption{Grouped Predicates by Meta-Feature Category and Meta-Feature Name across AutoClustering Frameworks}
\label{tab:grouped_predicates_sideways}
\begin{tabular}{p{3cm} p{3cm} >{\raggedright\arraybackslash}p{11cm} p{3cm}}
\toprule
\textbf{MF Category} & \textbf{MF} & \textbf{Predicate(s)} & \textbf{Framework(s)} \\
\midrule
\multirow{10}{*}{Landmarkers} 
    & \texttt{SIL} & \texttt{SIL > -0.24}, \texttt{SIL <= -0.18}, \texttt{SIL > -1.06} & autoclust, poac \\
    & \texttt{dbi} & \texttt{dbi <= -1.38} & autoclust \\
    & \texttt{ch} & \texttt{ch <= -2.63}, \texttt{ch >= -2.63} & autoclust \\
    & \texttt{cop} & \texttt{cop <= -0.36}, \texttt{cop <= -3.06} & autoclust \\
    & \texttt{di} & \texttt{di > -0.44}, \texttt{di <= -3.9} & autoclust \\
    & \texttt{DBS} & \texttt{DBS > 2.67} & poac \\
    & \texttt{reachability} & \texttt{reachability >= -7.51} & autocluster \\
    & \texttt{cohesiveness.mean} & \texttt{cohesiveness.mean > 3.61}, \texttt{cohesiveness.mean <= 577.58} & poac \\
    & \texttt{cji} & \texttt{cji <= -1.36} & autoclust \\
    & \texttt{t\_mean.sd} & \texttt{t\_mean.sd > -0.11}, \texttt{t\_mean.sd > 0.06}, \texttt{t\_mean.sd > 0.66} & poac, ml2dac \\
\midrule
\multirow{14}{*}{Statistical} 
    & \texttt{iq\_range} & \texttt{iq\_range.sd > 0.78}, \texttt{iq\_range.mean > 0.12}, \texttt{iq\_range.mean <= 0.12} & autocluster, poac \\
    & \texttt{mad.mean} & \texttt{mad.mean > 1.95}, \texttt{mad.mean > 3.7}, \texttt{mad.mean > 5.68} & autocluster, ml2dac \\
    & \texttt{mean.sd} & \texttt{mean.sd > 1.06}, \texttt{mean.sd > 0.52} & autocluster, ml2dac \\
    & \texttt{eigenvalues.mean} & \texttt{eigenvalues.mean >= 14.07}, \texttt{eigenvalues.mean > 3.64}, \texttt{eigenvalues.mean > 34.64} & autocluster \\
    & \texttt{inst\_to\_attr} & \texttt{inst\_to\_attr >= 4.16}, \texttt{inst\_to\_attr <= 6.0}, \texttt{inst\_to\_attr <= 416.66} & autocluster, poac \\
    & \texttt{min.mean} & \texttt{min.mean > -2.0}, \texttt{min.mean > -12.96} & autocluster \\
    & \texttt{conf.sd} & \texttt{conf.sd > 0.44} & autocluster \\
    & \texttt{cor.sd} & \texttt{cor.sd > 0.04} & autocluster \\
    & \texttt{var.sd} & \texttt{var.sd > 2.27}, \texttt{var.sd > 4.44} & ml2dac \\
    & \texttt{sd.sd} & \texttt{sd.sd > 0.44}, \texttt{sd.sd > 0.31} & ml2dac \\
    & \texttt{median.sd} & \texttt{median.sd > 7.0}, \texttt{median.sd > 1.37} & ml2dac \\
    & \texttt{cov.sd} & \texttt{cov.sd > 1.38} & ml2dac \\
\midrule
\multirow{3}{*}{Simple} 
    & \texttt{nr\_num} & \texttt{nr\_num > 6} & autocluster, ml2dac \\
    & \texttt{nr\_attr} & \texttt{nr\_attr <= 61.0} & poac \\
\midrule
\multirow{2}{*}{Info-Theoretical} 
    & \texttt{attr\_conc.sd} & \texttt{attr\_conc.sd > 2.27}, \texttt{attr\_conc.sd > 0.0} & ml2dac \\
    & \texttt{wg\_dist.sd} & \texttt{wg\_dist.sd <= -0.23}, \texttt{wg\_dist.sd <= 0.23} & poac \\
\bottomrule
\end{tabular}
\end{sidewaystable}

\end{document}